\documentclass[runningheads]{llncs}

\usepackage{eccv}

\usepackage{eccvabbrv}

\usepackage{graphicx}
\usepackage{booktabs}

\usepackage[accsupp]{axessibility}  

\usepackage{xcolor}
\usepackage{subcaption}
\usepackage{listings}
\usepackage{marvosym}
\usepackage{wrapfig}
\usepackage[numbers]{natbib}
\usepackage[normalem]{ulem}

\usepackage{xr-hyper}

\usepackage[pagebackref,breaklinks,colorlinks,allcolors=cvprblue]{hyperref}
\usepackage[capitalize,noabbrev]{cleveref}

\crefname{fig.}{Fig.}{Figs.}
\Crefname{figure}{Figure}{Figures}
\crefname{tab.}{Tab.}{Tabs.}
\Crefname{table}{Table}{Tables}
\crefname{eq.}{Eq.}{Eqs.}
\Crefname{equation}{Equation}{Equations}
\crefname{sec.}{Sec.}{Secs.}
\Crefname{section}{Section}{Sections}

\definecolor{cvprblue}{rgb}{0.21,0.49,0.74}

\usepackage{orcidlink}
\usepackage{arydshln}

\begin{document}

\title{Training-Free Multi-Concept Image Editing} 

\titlerunning{Training-Free Multi-Concept Image Editing}


\author{Niki Foteinopoulou\inst{1}\orcidlink{0000-0003-4481-9360} \and
Ignas Budvytis\inst{2}\orcidlink{0000-0002-4278-198X} \and
Stephan Liwicki\inst{1}\orcidlink{0000-0002-2002-3627}}

\authorrunning{N. Foteinopoulou et al.}

\institute{Cambridge Research Laboratory, Toshiba Europe, Cambridge, UK \\
\email{\{niki.foteinopoulou, stephan.liwicki\}@toshiba.eu} \and
Independent Researcher, UK \\
\email{ignas.budvytis@gmail.com}}

\maketitle

\begin{abstract}
Training-free image editing with diffusion models is highly desirable yet is complex and remains a significant challenge. While recent optimisation-based methods achieve strong zero-shot edits from text, they still struggle to preserve identity and capture intricate details, such as facial structure, surface texture, or object-specific geometry, that exist below the level of linguistic abstraction. To address this fundamental gap, we propose Concept Distillation Sampling (CDS). To the best of our knowledge, we are the first to introduce a unified, training-free framework for target-less, multi-concept image editing. 

CDS overcomes this linguistic bottleneck of previous methods by anchoring the editing process in the certainty of pretrained LoRA adapters. We integrate a highly stable distillation backbone (featuring ordered timesteps, regularisation, and negative-prompt guidance) with a novel dynamic weighting mechanism. This approach enables the composition and control of multiple visual concepts directly within the diffusion process, utilising spatially-aware priors from pretrained LoRA adapters without causing concept clashing. Our method preserves instance concept identity without requiring reference samples of the desired edit. Extensive quantitative and qualitative evaluations demonstrate that CDS establishes a new state-of-the-art over existing training-free editing and multi-LoRA composition methods on the InstructPix2Pix and ComposLoRA benchmarks. Project Page: \url{https://nickyfot.github.io/cds/}.
\end{abstract}    

\section{Introduction}
\label{sec:intro}

Recent advances in conditional image generation have brought us remarkably close to human-level quality in both image creation and editing~\cite{rombach_high-resolution_2022, Ramesh_2022_DALLE2, Saharia_2022_NeurIPS, labs2025flux1kontextflowmatching}. This progress is largely driven by diffusion models trained on vast, diverse datasets~\cite{schuhmann2022laion5b, wang2022diffusiondb}, which enable high-fidelity synthesis guided by flexible conditioning signals such as text, sketches, or images. Among these, text-to-image diffusion models have become particularly influential, thanks to their intuitive natural-language control, which often allows high-level control without requiring task-specific fine-tuning. Training-free approaches are particularly valuable in this context, as they bypass the steep computational costs and data-collection burdens of per-image or per-task fine-tuning. Furthermore, as foundational diffusion models continue to scale and improve, training-free methods inherently inherit their enhanced zero-shot capabilities, offering a highly adaptable and accessible solution for complex image manipulation.

\begin{figure}[t]
\centering
\includegraphics[width=\linewidth]{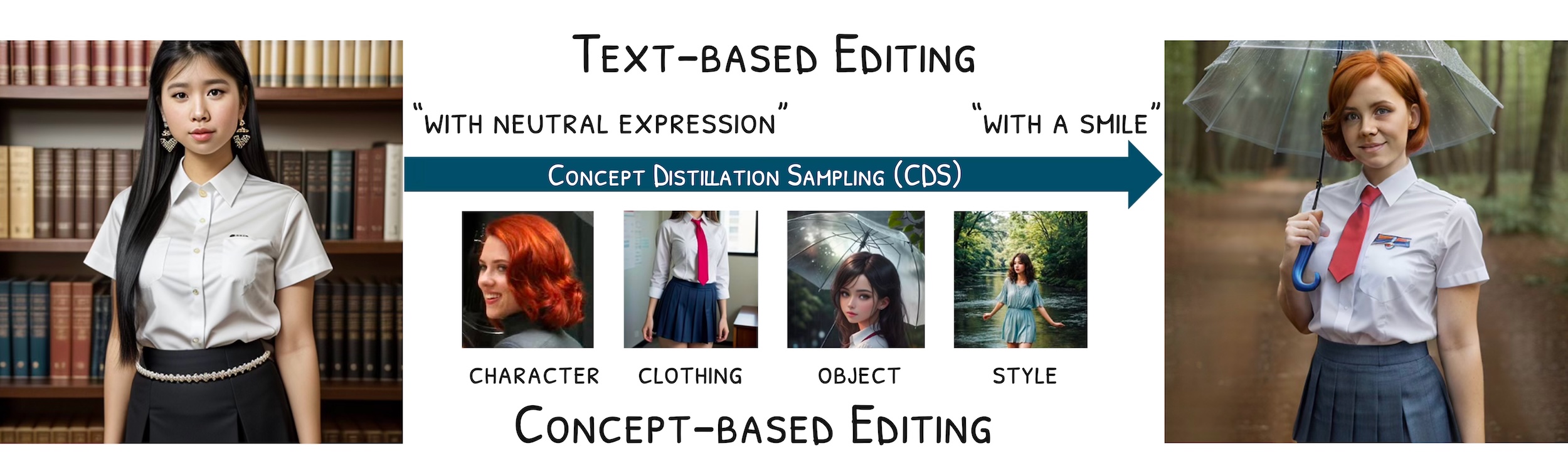}
\caption{Our method enables training-free, concept-based image editing with diffusion models. While recent state-of-the-art optimisation methods rely solely on text guidance and often lose instance identity during complex, multi-element changes, we unify a regularised optimisation objective with LoRA-driven concept composition. This enables multiple semantic or visual concepts (\eg facial structure, clothing, objects, or backgrounds captured in Dreambooth-style LoRAs) to be combined and controlled directly within the diffusion process. This approach preserves concept identity and fine-grained details that would be impossible to adequately describe in text alone, bridging the gap between text-based and visual concept-driven control.}\label{fig:teaser}
\end{figure}
However, editing existing images poses a subtler challenge than generating them from scratch. Effective editing requires not only semantic understanding but also content consistency, ensuring that the edited image remains recognisably the same subject. Recent training-free optimisation-based methods such as Delta Denoising Score (DDS)~\cite{Hertz_2023_ICCV} achieve impressive zero-shot edits by refining the Score Distillation Sampling (SDS) objective~\cite{pooledreamfusion}, providing strong semantic control when an edit can be clearly expressed in natural language. Yet, language alone is an insufficient descriptor for all visual concepts. Many attributes defining identity, such as structure, texture, or object-specific geometry, exist below the level of linguistic abstraction (\eg, the exact curvature of a specific character's jawline, or the unique wear patterns on a specific garment). As a result, state-of-the-art text-only methods often fail when edits involve multiple entities or concept-specific features that cannot be captured purely by text prompts.

In parallel, personalisation methods such as DreamBooth~\cite{ruiz2023dreambooth} and Low-Rank Adaptation (LoRA)~\cite{hu2022lora} have demonstrated how instance-specific information can be encoded directly into a diffusion model’s parameters. LoRAs effectively serve as concept representations, embedding appearance, style, or identity traits that text cannot describe. However, current multi-LoRA composition techniques~\cite{zhongmulti2024, meral2024clora, foteinopoulou2025loratorio} are primarily designed for text-to-image generation, not for image editing, where spatial alignment and subject consistency must be maintained.

To bridge this gap, we propose Concept Distillation Sampling (CDS), a novel method for preserving identity in multi-LoRA image editing. Our work unifies optimisation-based image editing and LoRA-based concept composition into a single framework for concept-aware, instance-consistent editing. DDS offers precise control via optimisation, while LoRAs provide low-level, semantically grounded priors that anchor edits to specific concepts or instances. Combining the two enables edits that are both semantically meaningful and visually consistent across multiple concepts (\cref{fig:teaser}). Here, a concept denotes any coherent visual attribute encoded by a LoRA adapter that text alone cannot capture. Each LoRA thus serves as a latent prior and low-level anchor within the optimisation loop (\cref{fig:method}), enabling controlled, consistent multi-concept editing.

To the best of our knowledge, we are the first to introduce a unified, training-free framework that combines the tasks of multi-LoRA composition and image editing. Unlike text-only SOTA methods~\cite{Koo:2024PDS, kim2025dreamcatalyst} that suffer from gradient ambiguity during complex edits, CDS effectively overcomes the linguistic bottleneck of previous distillation methods by optimising and anchoring the editing process in the certainty of the LoRA adapters. Our proposed CDS framework comprises two synergistic components:
First, we propose an optimised improvement upon previous distillation sampling works for 2D text-guided image editing. By revisiting DDS, we identify and resolve key design factors that limit stability and edit fidelity through principled refinements in timestep ordering, regularisation, and negative-prompt guidance. Second, we introduce a mechanism that dynamically weighs the contribution of each concept-LoRA. This mechanism performs patch-wise weighting of multiple LoRA outputs based on their feature similarity with the base model’s output, allowing multiple concepts to be seamlessly composed and controlled directly within the diffusion process (\cref{fig:teaser,fig:method}).

Furthermore, our method is truly generalisable. Any concept-LoRA can be incorporated into the editing process without requiring reference samples of the desired final edit. Recent works~\cite{manor2026spanningvisualanalogyspace} explore the idea of concept-based image editing; however, requiring target images for novel concept composition is inherently counter-intuitive to the goal of creating entirely novel, out-of-distribution compositions (\ie, unique, synthetic edits that combine elements in ways not present in any existing source image)s. CDS bypasses this limitation entirely, yielding stable, spatially-aware, and concept-preserving edits under a strict training-free constraint.
Our contributions are summarised as follows:

\begin{enumerate}
    \item We propose Concept Distillation Sampling (CDS), the first unified, training-free framework that formalises a new task: zero-shot, reference-free, multi-concept editing, and provides the first unified framework that addresses it.

    \item The first component of CDS is a refined delta-denoising formulation that improves stability and fidelity in zero-shot editing, coupled with a novel dynamic weighting method that patch-wise balances the contribution of multiple concept-LoRAs without retraining.

    \item We are the first to formalise the combined challenge of zero-shot multi-LoRA composition and training-free image editing. Furthermore, we show that CDS generalises across diverse editing scenarios. From text-driven zero-shot edits to complex pose transformations in InstructPix2Pix~\cite{brooks2022instructpix2pix}, and multi-element composition on ComposLoRA~\cite{zhongmulti2024}, CDS achieves state-of-the-art results, proving vastly superior to naive adaptations of existing text-to-image composition methods (\eg, strategies like Switch or Merge applied to editing).

\end{enumerate}

\section{Related Work}
\label{sec:related}
\textbf{Diffusion-based Editing:} Recent diffusion-based editing methods adapt text-to-image models for image modification by exploiting semantic priors. Approaches like Prompt-to-Prompt~\cite{mokady2022null}, DiffEdit~\cite{couairon2022diffedit}, and MasaCtrl~\cite{cao2023masactrl} use attention manipulation, masking, or diffusion inversion for localised, text-guided edits, while Null-Text~\cite{mokady2022null} enhances reconstruction fidelity. However, their reliance on text conditioning often compromises instance preservation across edits. Optimisation-based methods such as SDS~\cite{pooledreamfusion} align parametric representations with diffusion priors but can cause unwanted global changes. DDS~\cite{Hertz_2023_ICCV} mitigates this via differential gradients, and PDS~\cite{Koo:2024PDS} adds regularisation to retain image composition (\cref{sec:preliminaries}). Recent work like DreamCatalyst~\cite{kim2025dreamcatalyst} further improves identity preservation, though it remains inherently restricted to 3D scene editing. These methods randomly sample steps and remain confined strictly to text-to-image edits.

\textbf{Composable Text-to-Image Generation} seeks to synthesise images that coherently blend several concepts. Early work used structured conditioning (layouts or scene graphs) to improve compositionality~\citep{johnson2018image, song2021scorebased, gafni2022make}. Diffusion-based methods later refined the generative process or prompting to handle multi-concept inputs~\citep{feng2023trainingfree, huang2023composer, kumari2023multi, lin2023designbench, ouyang2025k}, but often depend on prompt design. Modular systems~\citep{du2020compositional, liu2021learning, li2023composing, simsar2025loraclr} combine separately trained models yet require heavy fine-tuning. We instead propose CDS, a unified training-free framework that dynamically composes LoRA adapters directly into the image editing process.

\textbf{LoRA-Based Skill Composition}. LoRA provides lightweight personalisation for large diffusion models \citep{ruiz2023dreambooth, sohn2023styledrop} in image generation. Approaches such as LoRAHub, ZipLoRA, and related mixture or hypernetwork-based schemes~\citep{huang2023lorahub, shah2024ziplora, zhu_mole_2024, shenaj_lorarar_2024, ruiz2024hyperdreambooth} combine multiple LoRAs through learned weighting, but require supervision and offer limited generalisation. Arithmetic or attention-driven merging (LoRA Merge, CLoRA) \citep{huggingface2024merge, meral2024clora} and inference-time strategies (LoRA Switch, Composite, MultiLFG) \citep{zhongmulti2024, zou_cached_2025, roy2025multlfg} avoid retraining yet suffer from instability and inefficiency as LoRAs accumulate. LoRAtorio~\cite{foteinopoulou2025loratorio} proposes using the distance of LoRA-enhanced model predictions from that of the base model as a proxy of certainty. Concurrent works~\citep{manor2026spanningvisualanalogyspace} target concept-based images with source-target example image pairs. In contrast, we enable multi-concept editing intuitively, which will synthesise novel views without explicit guidance. None of the existing multi-LoRA composition methods is formulated for training-free, target-less image editing. \textbf{To our knowledge, CDS is the first framework to formalise the multi-concept editing task, allowing seamless and controllable integration of multiple concepts without requiring reference images of the desired edit.}
\section{Preliminaries}
\label{sec:preliminaries}

\textbf{Score Distillation Sampling (SDS):}
Score Distillation Sampling (SDS)~\cite{pooledreamfusion} leverages pretrained diffusion models to optimise a parametric image or 3D representation $x_\theta$, where $\theta$ denotes the optimisable image parameters. This is achieved by matching the model's predicted noise to that of a frozen teacher network $\epsilon_\phi(z_t, y, t)$ conditioned on a text prompt $y$. 
The SDS gradient can be written as:
\begin{equation}
\nabla_\theta \mathcal{L}_{\mathrm{SDS}}
= 
\mathbb{E}_{t,\epsilon}
\!\left[
w(t)\, 
\big(
\epsilon_\phi(z_t, y, t) - \epsilon
\big)
\frac{\partial z_0}{\partial \theta}
\right],
\end{equation}
where $w(t)$ is a time-dependent weighting term, $\epsilon \sim \mathcal{N}(0, \mathbf{I})$ is the injected noise, and $z_t$ denotes the noisy latent at timestep $t$. Here, $z_0$ represents the clean latent encoding of the generated image, making $\frac{\partial z_0}{\partial \theta}$ the Jacobian of the differentiable encoding and rendering process. 
The timesteps $t$ are uniformly randomly sampled, ensuring denoising across varying noise magnitudes rather than following the forward diffusion order.
Although SDS provides an effective means for optimising image parameters under text guidance, it often introduces undesired global updates when used for editing (\cref{fig:optimiseddds}).

\textbf{Delta Denoising Score (DDS):}
To mitigate the spurious gradients of SDS in image editing, Delta Denoising Score (DDS)~\cite{Hertz_2023_ICCV} removes the source-conditioned denoising component from the target's score estimate, leaving only the differential (``delta'') signal corresponding to the semantic change between prompts. 
The DDS gradient is given by:
\begin{equation}
\nabla_\theta \mathcal{L}_{\mathrm{DDS}}
= \mathbb{E}_{t, \epsilon}\Big[
w(t)\,
\big(
\epsilon_\phi(z_t^{\mathrm{tgt}}, y_{\mathrm{tgt}}, t)  -
\epsilon_\phi(z_t^{\mathrm{src}}, y_{\mathrm{src}}, t)
\big)
\Big]
\frac{\partial z^{\mathrm{tgt}}_0}{\partial \theta},
\label{eq:dds}
\end{equation}
where $z_t^{\mathrm{src}}$ and $z_t^{\mathrm{tgt}}$ denote the noised versions of the source and target latents, respectively, while $y_{\mathrm{src}}$ and $y_{\mathrm{tgt}}$ are their associated prompts.  
When the prompts coincide ($y_{\mathrm{tgt}}{=}\,y_{\mathrm{src}}$), the DDS gradient vanishes, thus preserving unedited regions. 
This formulation leads to more localised targeted edits with reduced blurring, although unintentional environment edits are not entirely eradicated.

To further constrain the distillation process, PDS~\cite{Koo:2024PDS} and DreamCatalyst~\cite{kim2025dreamcatalyst} in 3D/NeRF editing explore aligning the posterior trajectories between source and target domains. In practice, this introduces a regularisation term scaling with the difference between the source and target latents ($\hat{\epsilon}_t^{\mathrm{tgt}} - \hat{\epsilon}_t^{\mathrm{src}}$ or $z_0^{\mathrm{tgt}} - z_0^{\mathrm{src}}$). These formulations rely on time-dependent weighting functions derived strictly from the base model's variance schedule and vanish when ordering timesteps~\cite{Koo:2024PDS}.

\begin{figure}[t]
    \centering
    \includegraphics[width=\linewidth]{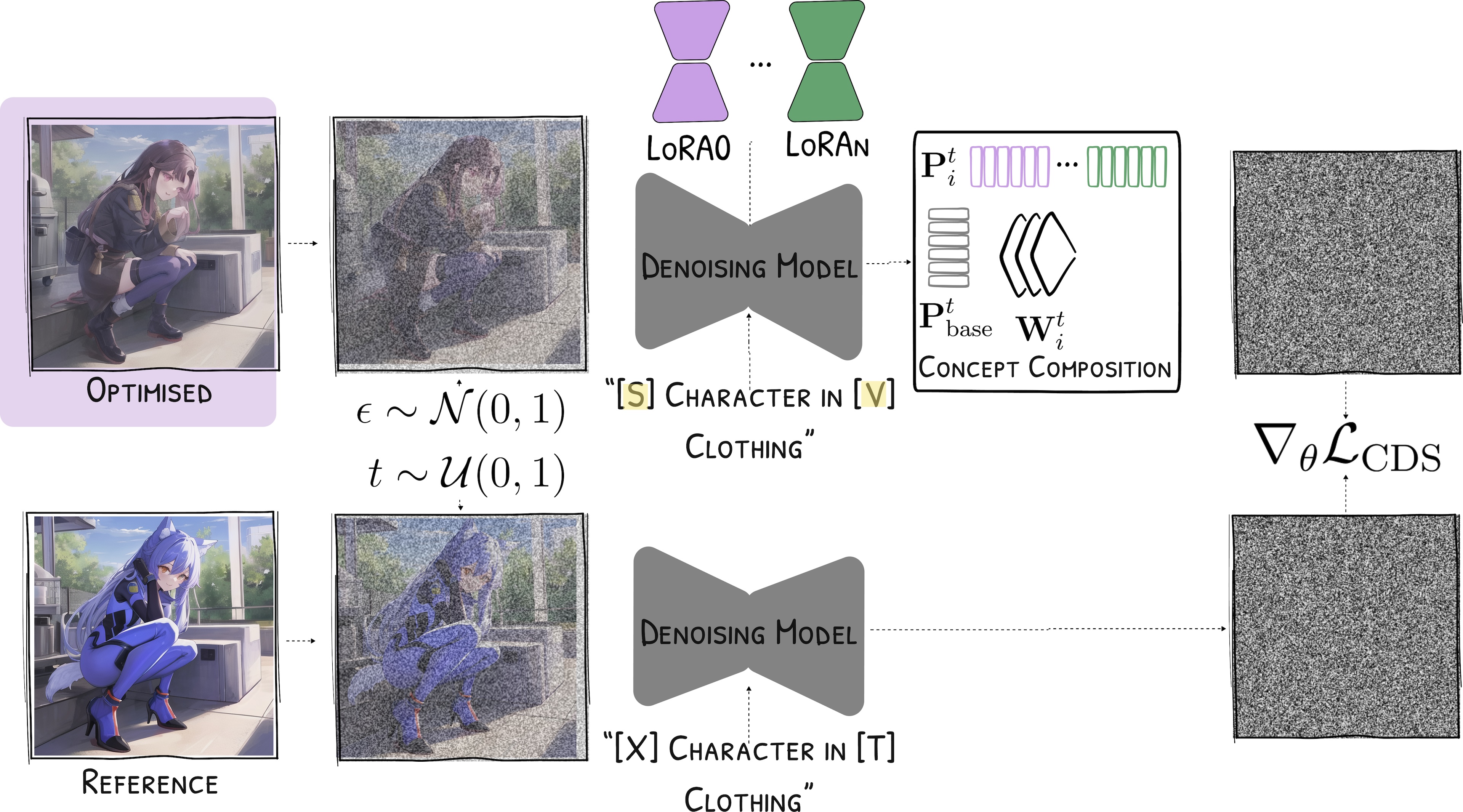}
    \caption{\textbf{Overview of our CDS image editing framework.} We integrate multiple LoRA adapters into the image optimisation loop, enabling spatially-aware, concept-preserving edits across multiple concepts. The denoising trajectory follows ordered timesteps, allowing precise control over pose, style, and semantic attributes while maintaining overall composition.}
    \label{fig:method}
\end{figure}

\section{Concept Distillation Sampling}
\label{sec:method}
In this section, we formalise the problem of target-less, multi-concept image editing. To solve this, we introduce Concept Distillation Sampling, a unified training-free framework. CDS comprises two core innovations: a regularised, timestep-ordered distillation objective that guarantees structural stability, and a dynamic concept-weighting mechanism that seamlessly composes multiple LoRA adapters without spatial interference. An overview is provided in \cref{fig:method}.

\begin{figure*}[t]
    \centering
    \includegraphics[width=\linewidth]{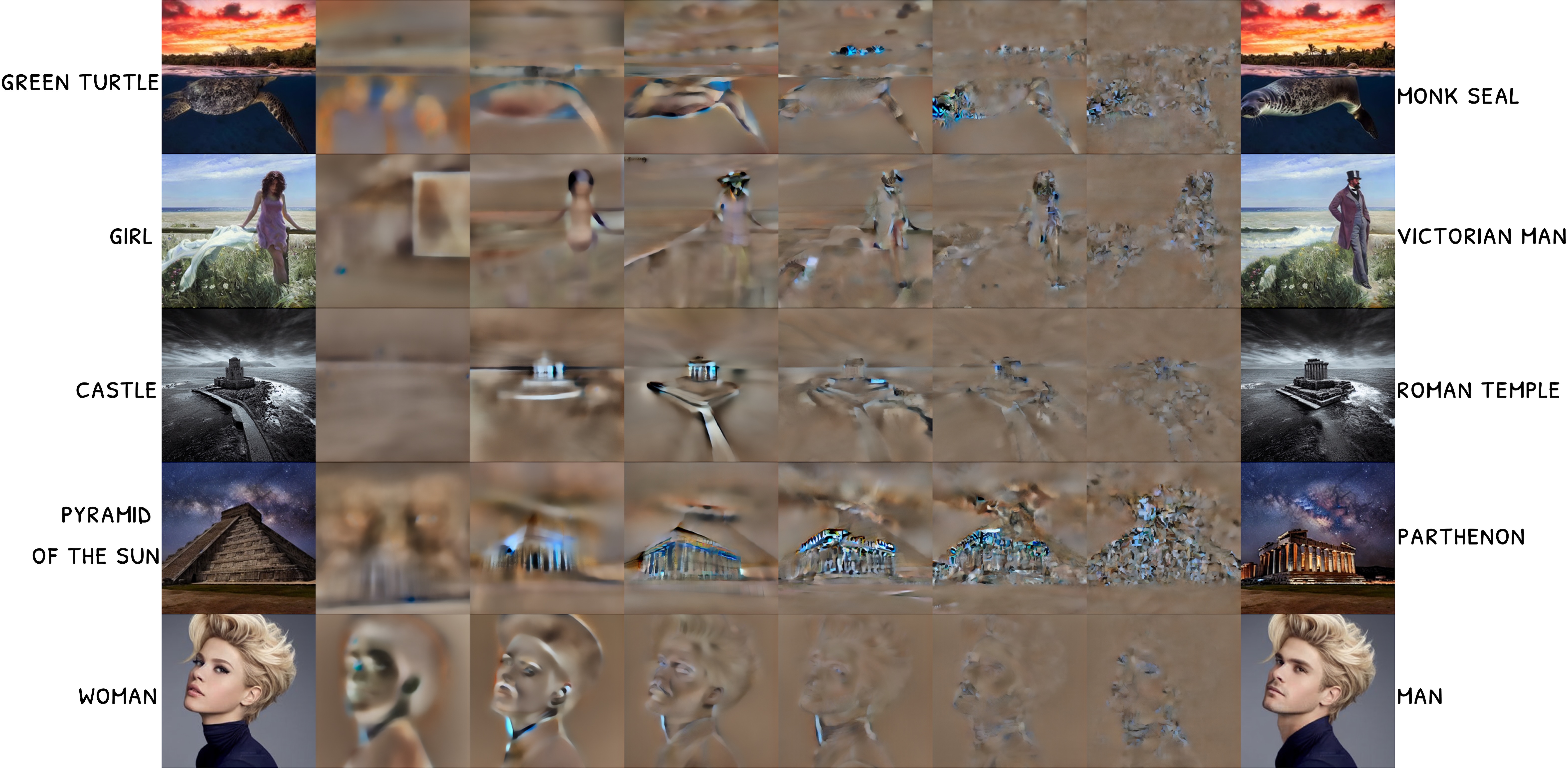}
    \caption{\textbf{Visualisation of timestep-ordered denoising.} Comparison between the source image, $\nabla_\theta \mathcal{L}_{\mathrm{CDS}}$ at intermediate optimisation steps, and the final target image. Unlike DDS~\cite{Hertz_2023_ICCV}, which samples timesteps uniformly at random, our method enforces a strict descending timestep order, enabling a coarse-to-fine denoising trajectory as is evident from the visualised gradients. Early steps capture high-frequency structural details such as edges, while later steps refine lower-frequency and stylistic components, resulting in a more coherent and stable edit.}
    \label{fig:denoising}
\end{figure*}

\textbf{Problem Formulation:}
\label{sec:problem_formulation}
We address the task of zero-shot concept-based image editing.  In this context, a ``concept'' refers to any semantically meaningful attribute that can be encoded via a LoRA adapter (\eg a specific character identity, object geometry, or artistic style) that is encoded within a personalised LoRA adapter, but which cannot be adequately isolated or described by natural language alone. Each concept is assumed to be learned independently, to reflect truly in-the-wild zero-shot settings.

Given a source image $x_{\mathrm{src}}$, a target text prompt $y^{\mathrm{tgt}}$, and a set of $N$ concept-encoding LoRA modules $\{lora_1, lora_2, \dots, lora_N\}$, our objective is to synthesise a target image $x_{\mathrm{tgt}}$. This image must preserve the unedited structural foundations of $x_{\mathrm{src}}$ while seamlessly integrating the non-linguistic concepts dictated by the LoRAs. We enforce a strict zero-shot constraint, that is, no reference samples of the desired final edit are available to guide the spatial composition. To achieve this, the editing process must satisfy two conditions simultaneously. First, the parametrised target $x_0^{\mathrm{tgt}}$ must be optimised such that the differential noise aligns with the desired semantic shift:

\begin{equation}
    \epsilon_{\mathrm{base}}(x_t^{\mathrm{tgt}}, y^{\mathrm{tgt}}, t) - \epsilon_{\mathrm{base}}(x_t^{\mathrm{src}}, y^{\mathrm{src}}, t) \approx x_0^{\mathrm{tgt}} - x_0^{\mathrm{src}}.
\end{equation}

Second, the standard base noise estimate $\epsilon_{\mathrm{base}}$ must be dynamically replaced by a spatially aware, composite noise prediction derived from the $N$ independent LoRA modules, preventing concept clash.

\subsection{CDS Optimisation Objective}~\label{sec:optimised_dds}

\begin{figure}[t]
    \centering
    \includegraphics[width=\linewidth]{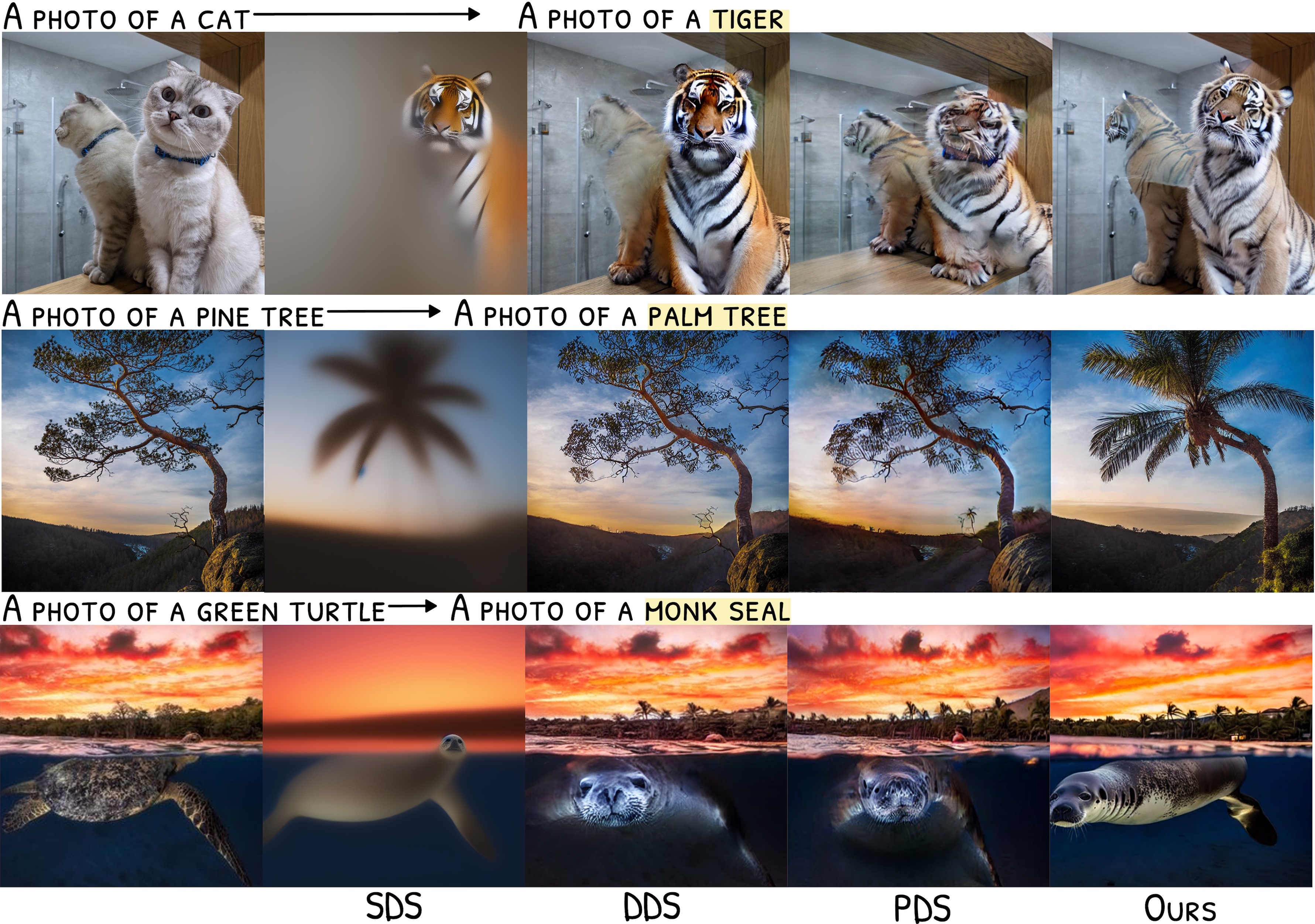}
    \caption{Comparison of our optimisation objective vs baselines. Our method produces better results for the target subject with similarly perceived changes in the background.}
    \label{fig:optimiseddds}
\end{figure}
Standard distillation objectives sample timesteps $t \sim \mathcal{U}(0,1)$ uniformly at random. While this promotes generalisation across noise levels, it fundamentally ignores the temporal structure of the diffusion reverse process, where early steps dictate high-frequency structural edges and later steps govern low-frequency stylistic rendering. To anchor the concept editing process, CDS enforces a strict descending timestep order, $1 > u > \dots > v > 0$. This forces a coarse-to-fine denoising trajectory (\cref{fig:denoising}). However, deterministic ordering introduces destabilising gradients if unconstrained. To counter this without suffering the vanishing gradients of prior trajectory alignment works, we propose a training schedule-independent regularisation objective:

\begin{equation}
    \nabla_\theta \mathcal{L}_{\mathrm{CDS}} = \mathbb{E}_{t, \epsilon_t, \epsilon_{t-1}}\Big[\big(\eta\,| x_0^{\mathrm{tgt}} - x_0^{\mathrm{src}}|+ (\hat{\epsilon}_t^{\mathrm{tgt}} - \hat{\epsilon}_t^{\mathrm{src}})\big)\frac{\partial x^{\mathrm{tgt}}_0}{\partial \theta}\Big],
\end{equation}

\noindent where $\hat{\epsilon}_t^{\mathrm{src}}$ and $\hat{\epsilon}_t^{\mathrm{tgt}}$ are the respective predicted noises, and $\eta$ is a static hyperparameter governing the strength of latent alignment. This explicit formulation prevents coefficients from decaying during sequential stepping, as opposed to~\cite{Koo:2024PDS} where ordered timesteps result to a zero derivative and thus no edit. To further guide the optimisation against degenerate visual modes commonly induced by aggressive LoRA conditioning, we integrate negative prompt guidance directly into the optimisation loop:
\begin{equation}
    \hat{\epsilon}_{\mathrm{guided}}(z_t, y^{+}, y^{-}, t) = (1 + \lambda) \epsilon(z_t, y^{+}, t) - \lambda \epsilon(z_t, y^{-}, t),
\end{equation}
\noindent where $\lambda$ dictates the classifier-free guidance scale. Together, timestep ordering, explicit regularisation, and negative guidance form a highly stable distillation backbone capable of supporting complex conceptual shifts.

\subsection{Dynamic Concept Weighting}\label{sec:dynamic_weighting}

While the $\mathcal{L}_{\mathrm{CDS}}$ objective provides the vehicle for stable text-guided editing, the core of concept-based editing lies in composing multiple LoRA adapters without concept clashing. A na\"ive merge of LoRA weights often affects instance fidelity, especially as multiple LoRAs are composed~\cite{zhongmulti2024}, and creates severe spatial artefacts, even in the simpler image generation task. We propose a dynamic, inference-time weighting mechanism that assesses a proxy for the confidence of each concept at every denoising step. Our core intuition is that if a LoRA-enhanced model predicts noise highly similar to the unmodified base model, that LoRA is not contributing meaningful concept-specific information. 
On the contrary, where the prediction diverges strongly from the base model, the LoRA is actively injecting its encoded concept. 

Let $\epsilon_{\mathrm{base}}(z_t, t, c)$ be the noise prediction of the frozen base model, and the prediction of the $i$-th LoRA adapter be
$\epsilon_{lora_i}(z_t, t, c)$. At each timestep $t$, we extract the spatial feature maps and partition them into $P$ non-overlapping patches. We define a flattening function $\psi$ that maps these spatial features into a set of vectors:

\begin{equation}
    \mathbf{P}_{\mathrm{base}}^t = \psi(\epsilon_{\mathrm{base}}(z_t, t, c)), \quad \mathbf{P}_{i}^t = \psi(\epsilon_{lora_i}(z_t, t, c)).
\end{equation}

For every patch $p \in \{1, \dots, P\}$, we compute the cosine similarity between the base model's patch and the corresponding patch from the $i$-th LoRA:

\begin{equation}
    S_{i,p}^t = \langle \mathbf{P}_{\mathrm{base}, p}^t, \mathbf{P}_{i, p}^t \rangle_{\mathrm{cos}}.
\end{equation}
High similarity indicates low concept injection. Therefore, we compute the adaptive spatial weight $\omega_{i,p}^t$ for each LoRA patch by applying a temperature-scaled SoftMin operation across all $N$ concepts:

\begin{equation}
    \omega_{i,p}^t = \mathrm{SoftMin}_\tau(S_{i,p}^t) = \frac{\exp(-S_{i,p}^t / \tau)}{\sum_{j=1}^N \exp(-S_{j,p}^t / \tau)},
\end{equation}
\noindent where the temperature $\tau$ modulates the sharpness of the spatial isolation. These patch-wise weights are then upsampled back to the original spatial dimensions via nearest-neighbour interpolation, denoted as $\mathbf{W}_i^t$.The final, multi-concept conditional noise prediction used within the $\mathcal{L}_{\mathrm{CDS}}$ optimisation loop is thus constructed dynamically:

\begin{equation}
    \tilde{\epsilon}_{\mathrm{concept}}(z_t, t, c) = \sum_{i=1}^N \mathbf{W}_i^t \odot \epsilon_{lora_i}(z_t, t, c),
\end{equation}
\noindent where $\odot$ denotes the Hadamard (element-wise) product. By evaluating concept confidence iteratively and spatially, CDS ensures that distinct elements, such as a specific character's face from one LoRA and a unique clothing style from another, can seamlessly be composed in the edited image without concept confusion, while maintaining the source image structure.

\begin{table}[t]
\centering
\caption{\textbf{Quantitative Evaluation of our optimisation method against previous SoTA, on InstructPix2Pix.} Our method achieves statistically significant improvements in CLIPScore while maintaining comparable LPIPS; the CLIPScore gain reflects a genuine semantic edge under a deliberate semantic–fidelity trade-off.}
\label{tab:optdds}
\begin{tabular}{lcc}
\toprule
                & CLIPScore$\uparrow$      & LPIPS$\downarrow$       \\
\midrule
DiffusionClip$\ddag$   & 0.251 ± 0.022            & 0.572 ± 0.0594          \\
PnP$\ddag$             & 0.221 ± 0.036            & 0.310 ± 0.075           \\
InstructPix2Pix$\ddag$ & 0.219 ± 0.037            & 0.322 ± 0.215           \\
DDS$\ddag$             & 0.225 ± 0.031            & \uline{0.104 ± 0.061}  \\
PDS\Cross              & \uline{0.298 ± 0.044}    & \textbf{0.096 ± 0.033}                  \\
\textbf{Ours}          & \textbf{0.308 ± 0.042}*  & \uline{0.100 ±  0.042} \\
\bottomrule
\end{tabular}

\ddag Reported in~\cite{Hertz_2023_ICCV} \\
\Cross PDS~\cite{Koo:2024PDS} is a 3D method adapted to RGB\\
 * statistically significant at 99\% confidence interval.

\end{table}

\section{Results}
\label{sec:results}
 We summarise the core insights of our evaluation. First, we demonstrate that the foundational CDS distillation objective achieves state-of-the-art text-guided editing by finding a balance between semantic edit strength and structural preservation. Second, in the novel and highly challenging task of multi-concept editing (up to 5 LoRAs), our full CDS framework successfully prevents the severe concept erasure and spatial artefacts that plague existing composition baselines. Finally, comprehensive human and GPT-4V evaluations validate that CDS yields vastly superior visual harmony and instance fidelity, confirming that our spatially-aware dynamic weighting mechanism successfully bridges the gap between text-driven and concept-driven control.

\begin{table*}[t]
\centering
\caption{
    \textbf{Quantitative Evaluation of Concept Distillation Sampling (CDS).} \\
    Performance comparison across four LoRA composition strategies for different numbers of LoRAs on \textit{ComposLoRA} testbed. \\
    \textbf{Bold} values indicate statistical significance at the 99\% confidence interval. 
}
\label{tab:clipscore}
\begin{subtable}[t]{\textwidth}
\resizebox{\textwidth}{!}{%
\begin{tabular}{lllllllllll}
\hline
                   & \multicolumn{2}{c}{\textbf{$N=2$}}                                                & \multicolumn{2}{c}{\textbf{$N=3$}}                                                & \multicolumn{2}{c}{\textbf{$N=4$}}                                                & \multicolumn{2}{c}{\textbf{$N=5$}}                                                & \multicolumn{2}{l}{Total}                                                       \\ \cline{2-11} 
                   & \multicolumn{1}{c}{CLIPScore$\uparrow$} & \multicolumn{1}{c}{LPIPS$\downarrow$} & \multicolumn{1}{c}{CLIPScore$\uparrow$} & \multicolumn{1}{c}{LPIPS$\downarrow$} & \multicolumn{1}{c}{CLIPScore$\uparrow$} & \multicolumn{1}{c}{LPIPS$\downarrow$} & \multicolumn{1}{c}{CLIPScore$\uparrow$} & \multicolumn{1}{c}{LPIPS$\downarrow$} & \multicolumn{1}{c}{CLIPScore$\uparrow$} & \multicolumn{1}{c}{LPIPS$\downarrow$} \\ \hline
\textbf{CDS}       & 0.340 $\pm$ 0.030                       & \textbf{0.401 $\pm$ 0.174}            & \textbf{0.361 $\pm$ 0.022}              & \textbf{0.461 $\pm$ 0.158}            & \textbf{0.370 $\pm$ 0.019}              & 0.534 $\pm$ 0.104                     & \textbf{0.368 $\pm$ 0.059}              & 0.503 $\pm$ 0.082                     & \textbf{0.359 $\pm$ 0.032}              & \textbf{0.474 $\pm$ 0.129}            \\
\textbf{Composite} & \textbf{0.351 $\pm$ 0.021}              & 0.542 $\pm$ 0.150                     & \textbf{0.369 $\pm$ 0.019}              & 0.613 $\pm$ 0.129                     & \textbf{0.377 $\pm$ 0.018}              & 0.644 $\pm$ 0.077                     & 0.363 $\pm$ 0.057                       & 0.664 $\pm$ 0.039                     & \textbf{0.365 $\pm$ 0.028}              & 0.615 $\pm$ 0.098                     \\
\textbf{Switch}    & \textbf{0.350 $\pm$ 0.021}              & 0.456 $\pm$ 0.145                     & \textbf{0.361 $\pm$ 0.021}              & 0.485 $\pm$ 0.147                     & 0.367 $\pm$ 0.020                       & \textbf{0.467 $\pm$ 0.128}            & 0.363 $\pm$ 0.057                       & \textbf{0.487 $\pm$ 0.108}            & \textbf{0.360 $\pm$ 0.029}               & \textbf{0.473 $\pm$ 0.132}            \\
\textbf{Merge}     & 0.346 $\pm$ 0.016                       & 0.547 $\pm$ 0.133                     & 0.352 $\pm$ 0.018                      & 0.601 $\pm$ 0.112                     & 0.354 $\pm$ 0.021                       & 0.634 $\pm$ 0.067                     & 0.346 $\pm$ 0.061                       & 0.662 $\pm$ 0.038                     & 0.349 $\pm$ 0.029                       & 0.610 $\pm$ 0.087                      \\ \hline
\end{tabular}%
}
\caption{Reality}
\end{subtable}

\begin{subtable}[t]{\textwidth}
    \resizebox{\textwidth}{!}{%
\begin{tabular}{lllllllllll}
\hline
                   & \multicolumn{2}{c}{\textbf{$N=2$}}                                                & \multicolumn{2}{c}{\textbf{$N=3$}}                                                & \multicolumn{2}{c}{\textbf{$N=4$}}                                                & \multicolumn{2}{c}{\textbf{$N=5$}}                                                & \multicolumn{2}{l}{Total}                                                       \\ \cline{2-11} 
                   & \multicolumn{1}{c}{CLIPScore$\uparrow$} & \multicolumn{1}{c}{LPIPS$\downarrow$} & \multicolumn{1}{c}{CLIPScore$\uparrow$} & \multicolumn{1}{c}{LPIPS$\downarrow$} & \multicolumn{1}{c}{CLIPScore$\uparrow$} & \multicolumn{1}{c}{LPIPS$\downarrow$} & \multicolumn{1}{c}{CLIPScore$\uparrow$} & \multicolumn{1}{c}{LPIPS$\downarrow$} & \multicolumn{1}{c}{CLIPScore$\uparrow$} & \multicolumn{1}{c}{LPIPS$\downarrow$} \\ \hline
\textbf{CDS}  & 0.324 $\pm$ 0.026                       & \textbf{0.321 $\pm$ 0.096}            & 0.327 $\pm$ 0.027                       & \textbf{0.313 $\pm$ 0.084}            & 0.331 $\pm$ 0.027                       & \textbf{0.314 $\pm$ 0.079}            & 0.318 $\pm$ 0.029                       & \textbf{0.306 $\pm$ 0.067}            & 0.325 $\pm$ 0.027                       & \textbf{0.314 $\pm$ 0.082}            \\
\textbf{Composite} & 0.324 $\pm$ 0.029                       & 0.658 $\pm$ 0.057                     & 0.320 $\pm$ 0.031                       & 0.67 $\pm$ 0.068                      & 0.327 $\pm$ 0.026                       & 0.668 $\pm$ 0.058                     & 0.310 $\pm$ 0.022                        & 0.663 $\pm$ 0.039                     & 0.320 $\pm$ 0.027                        & 0.665 $\pm$ 0.055                     \\
\textbf{Switch}    & 0.331 $\pm$ 0.028                       & 0.378 $\pm$ 0.07                      & 0.335 $\pm$ 0.03                        & 0.368 $\pm$ 0.053                     & 0.339 $\pm$ 0.025                       & \textbf{0.316 $\pm$ 0.047}            & \textbf{0.331 $\pm$ 0.024}              & 0.319 $\pm$ 0.043                     & \textbf{0.334 $\pm$ 0.027}              & 0.345 $\pm$ 0.053                     \\
\textbf{Merge}     & \textbf{0.341 $\pm$ 0.026}              & 0.388 $\pm$ 0.091                     & \textbf{0.341 $\pm$ 0.03}               & 0.428 $\pm$ 0.091                     & \textbf{0.340 $\pm$ 0.026}               & 0.473 $\pm$ 0.094                     & \textbf{0.331 $\pm$ 0.022}              & 0.498 $\pm$ 0.084                     & \textbf{0.338 $\pm$ 0.026}              & 0.447 $\pm$ 0.090                      \\ \hline
\end{tabular}%
}
\caption{Anime}
\end{subtable}
\end{table*}

\subsection{Implementation Details}
 For a fair comparison with previous works, we select \textit{stable-diffusion-v1.5}~\citep{rombach_high-resolution_2022} as the backbone for all experiments. For experiments on InstructPix2Pix, we select 1000 images from the train set for all zero-shot text-guided comparisons of our optimisation objective against baseline DDS. We use a guidance scale \( s = 10 \), and image size \( 512 \times 512 \) and checkpoint ``runwayml/stable-diffusion-v1-5''.
 
For our multi-concept editing experiments, we build upon the setup of~\cite{zhongmulti2024}, using \textit{ComposLoRA} tests that include 22 pre-trained LoRAs spanning characters, clothing, styles, backgrounds, and objects, and modify from text-to-image generation to image-editing (details in~\cref{appendix:composlora}). We use the ``Realistic\_Vision\_V5.1'' and ``Counterfeit-V2.5'' checkpoints for realistic and anime-style images, respectively. For the realistic subset,  a guidance scale \( s = 7 \), and image size \( 512 \times 512 \); for the anime subset we use \( s = 10 \). DPM-Solver++~\citep{lu2022dpm} is used as the sampler, with all LoRAs scaled by a weight of $0.8$. We empirically set the temperature $\tau = 0.002$. For all experiments, we set the size of each patch to $2\times2$. For all experiments, we select 300 denoising steps and set $\eta=0.5$. Since our method operates at inference time, all experiments are run on a single RTX A6000 GPU. Results are averaged over three runs. We provide an extended sensitivity analysis in~\cref{app:sensitivity}.

\label{sec:gpt4v}
\begin{figure}[t]
    \centering
    \begin{subfigure}[b]{0.49\linewidth}
        \centering
        \includegraphics[width=0.98\linewidth]{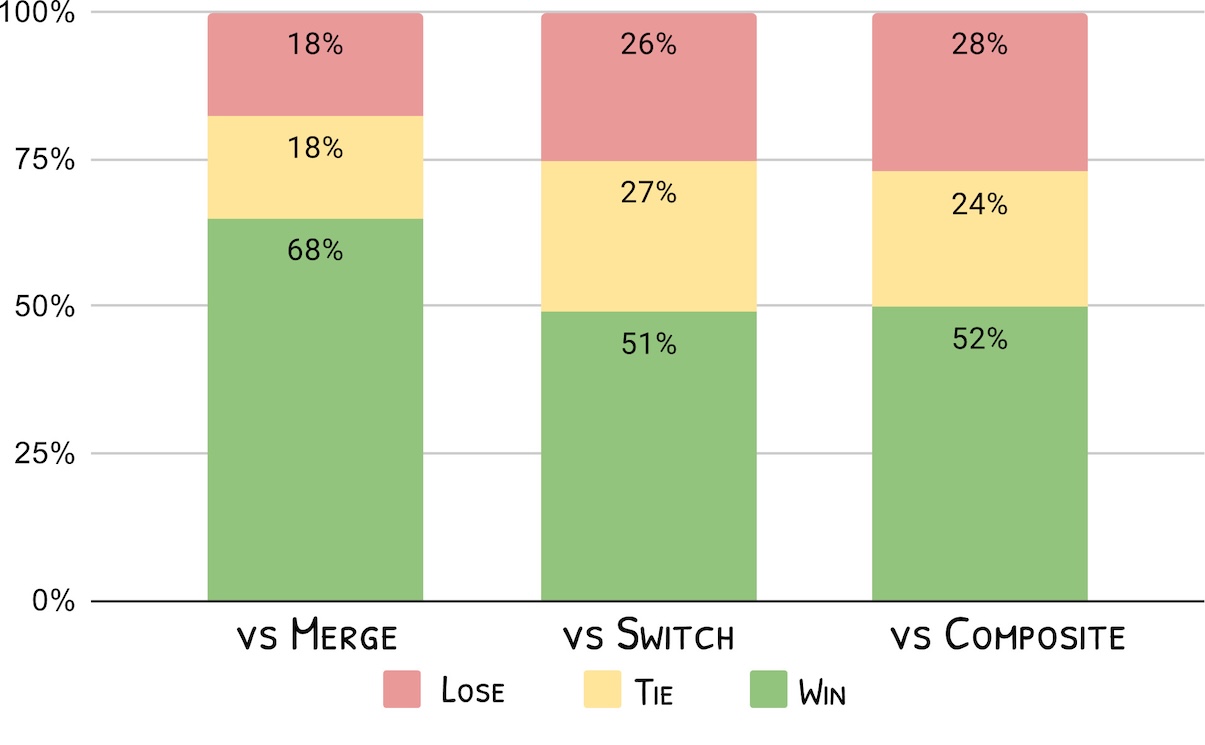}
        \caption{Win Rate (pairwise comparison).}
        \label{fig:winrate}
    \end{subfigure}
    \begin{subfigure}[b]{0.49\linewidth}
        \centering
        \includegraphics[width=0.98\linewidth]{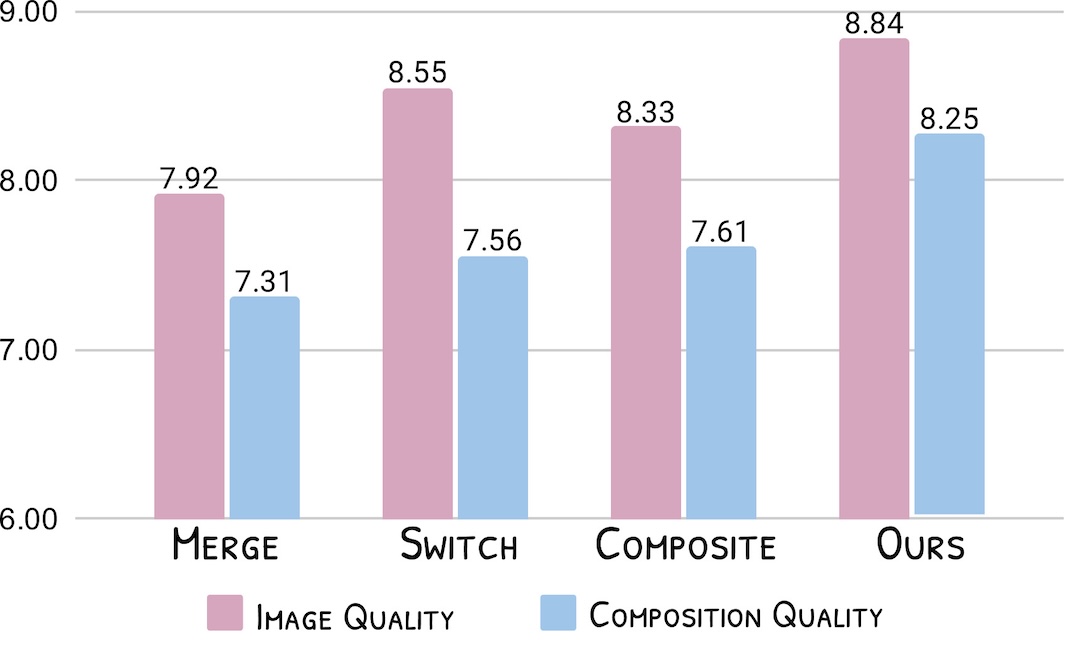}
        \caption{Image and Composition Quality.}
        \label{fig:gpt4}
    \end{subfigure}
    \caption{\textbf{Comparison of our CDS framework against previous LoRA-composition methods, evaluated using GPT-4V.} Our method achieves higher perceived image and composition quality, with consistently higher pairwise win rates.}
    \label{fig:gpt4_combined}
\end{figure}

\subsection{Quantitative Evaluation of Image Editing}

\textbf{Text-Guided Editing:} We first evaluate the core distillation objective of CDS on the InstructPix2Pix benchmark in~\cref{tab:optdds}. By isolating the optimisation backbone (ordered timesteps, regularisation, and negative prompts), we demonstrate its stability over prior distillation methods. Our method achieves a statistically significant improvement in CLIPScore over previous SoTA at 99\% confidence interval, while maintaining comparable LPIPS, visualised in~\cref{fig:optimiseddds}.
\noindent\textbf{Multi-Concept Editing (Full CDS):} To evaluate the effectiveness of our method in multi-concept editing, we evaluate the full CDS framework, which integrates our regularised distillation objective with dynamic concept weighting. We compare our proposed CDS against three existing multi-LoRA strategies: \textit{Composite}~\cite{zhongmulti2024}, \textit{Switch}~\cite{zhongmulti2024}, and \textit{Merge}~\cite{huggingface2024merge}, selected as they have publicly available code that can be adopted to image editing, without multiple inference steps.  
For each configuration, we generate scenes containing \(N\) distinct LoRAs, each representing a different concept from the \textit{ComposLoRA} testbed (character, clothing, object, background or style). We then apply our method to simultaneously edit all \(N\) entities into different but semantically consistent counterparts (\eg, character\(_1\) $\rightarrow$ character\(_2\), clothing\(_1\) $\rightarrow$ clothing\(_2\)). This process is repeated across all valid concept combinations in the testbed to ensure a comprehensive evaluation of both cross-concept consistency and edit controllability.

CDS achieves the lowest LPIPS across most configurations~\cref{tab:clipscore}; on the Reality total split, LPIPS is statistically tied with Switch, but CDS retains a clear advantage on Anime ($0.314$ vs. $0.345$) and on every per-N slice for $N \geq 3$.
CLIPScore remains on par or higher than baselines but saturates with more concepts, likely due to modality gaps and prompt ambiguity, suggesting that alone it's insufficient to capture perceptual and compositional quality in multi-concept edits. We see, however, that all methods have CLIPScores in similar ranges, which indicates that CLIPScore is adequate to measure whether the generic object is included, but not if the specific concept and instance are represented, thus highlighting the need for qualitative evaluation.

\subsection{Qualitative Evaluation via GPT-4V and Human Ranking}
Since CLIPScore alone cannot capture compositional quality, we complement it with qualitative evaluations using GPT-4V and human studies. 
GPT-4V scores and pairwise win rates against Switch, Composite, and Merge are shown in~\cref{fig:gpt4} and~\cref{fig:winrate}.
Human evaluators likewise ranked our method highest for image quality and concept integration, with the lowest average rank in~\cref{tab:human}, followed closely by Switch. Additional results can be seen in~\cref{app:qualitative}. 

\begin{figure*}[t]
    \centering
    \includegraphics[width=\linewidth]{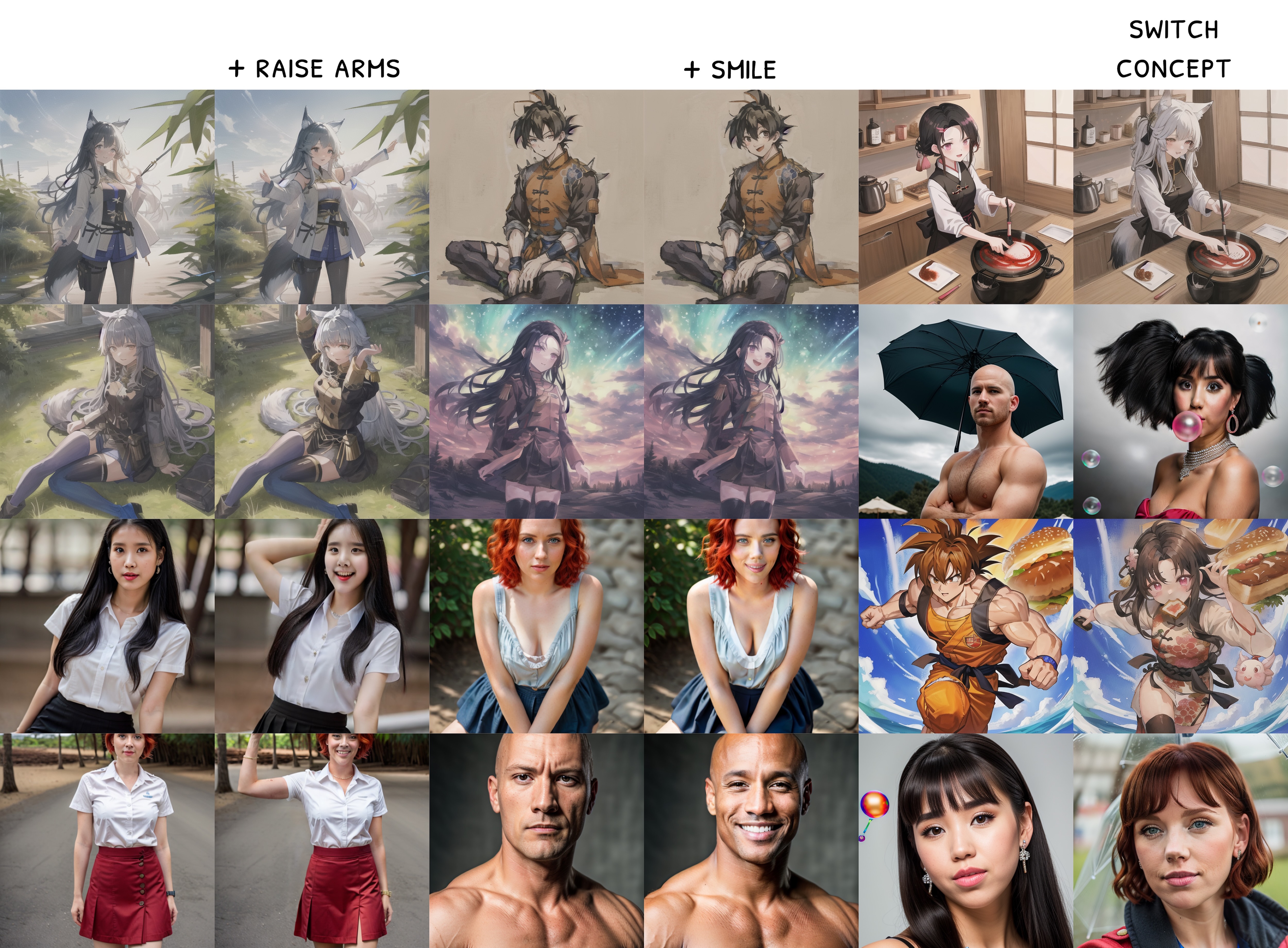}
    \caption{\textbf{Examples of our method under different editing conditions.}
    Our approach enables consistent and coherent edits across diverse scenarios, including pose changes, facial expression modifications, and element or object swaps (\eg, between different characters and environmental components). Each pair demonstrates the method’s ability to preserve subject and visual fidelity while adapting to the intended transformation.}
    \label{fig:samples}
\end{figure*}

\begin{table}[t]
\centering
\caption{\textbf{Human evaluation comparing our CDS framework to previous LoRA-composition methods.} In line with GPT-4V assessments, human raters consistently preferred our method, which achieved the lowest average rank and most frequent first-place selections (highest win rate).}
\label{tab:human}
\begin{tabular}{lcc}
\toprule
            &  Avg. Rank$\downarrow$        & Win Rate$\uparrow$                      \\
\midrule
    CDS     & \textbf{1.90}                 & \textbf{38\%}                           \\
    Compose & 2.62                          & 31\%                                    \\
    Switch  & 2.49                          & 25\%                                    \\
    Merge   & 3.00                          & 5\%                                     \\
\bottomrule
\end{tabular}
\end{table}

\subsection{Pose Changes Under Concept Preservation}
By unifying dynamic concept weighting with our distillation objective, CDS successfully handles complex edits involving simultaneous pose and semantic changes.
Shown in~\cref{fig:samples}, CDS maintains concepts across pose and expression variations while preserving key composition in multi-LoRA edits.

\subsection{Ablation}
\label{sec:ablation}
\begin{table}[t]
\centering
\caption{\textbf{CDS Core Objective Ablation.} We establish a baseline with standard DDS (300 steps unless indicated) and iteratively introduce our objective components to isolate their effects. Our unified formulation achieves the optimal balance between executing semantic changes and maintaining structural integrity.}
\label{tab:ddsablation}
\begin{tabular}{lcc}
\toprule
                     & CLIPScore$\uparrow$      & LPIPS$\downarrow$       \\
\midrule
DDS (200 steps)      & 0.225 ± 0.031            &  0.104 ± 0.061          \\
DDS                  & 0.305 ± 0.039            &  0.264 ± 0.061          \\\hdashline
Strict Timesteps (+DDS) & \textbf{0.332 ± 0.032}   &  0.461 ± 0.076          \\
Regularisation (+DDS)   & 0.291 ± 0.041            &  \textbf{0.081 ± 0.033} \\
Negative Prompts (+DDS)      & 0.318 ± 0.040            &  0.319 ± 0.076          \\
\textbf{Full CDS} & 0.308 ± 0.042            &  0.100 ±  0.042         \\
\bottomrule
\end{tabular}
\end{table}

As shown in~\cref{tab:ddsablation}, enforcing ordered timesteps increases both metrics, reflecting stronger structural changes consistent with more pronounced edits.
In contrast, adding regularisation lowers both metrics, indicating overly conservative optimisation that can suppress edits entirely. Strict Timesteps alone maximise CLIPScore ($0.332$) but at the cost of LPIPS ($0.461$); the L1 regulariser alone over-constrains the edit (CLIPScore $0.291$). The two work in cooperation; full CDS recovers structural fidelity (LPIPS $0.100$) without sacrificing semantic strength.
Balancing these competing forces within the CDS objective produces the best trade-off between edit strength and perceptual fidelity, aligning with empirical visual inspection. Additional ablations on the effect of chosen hyperparameters can be found in~\cref{app:sensitivity}.

\begin{figure}[t]
    \centering
    \includegraphics[width=\linewidth]{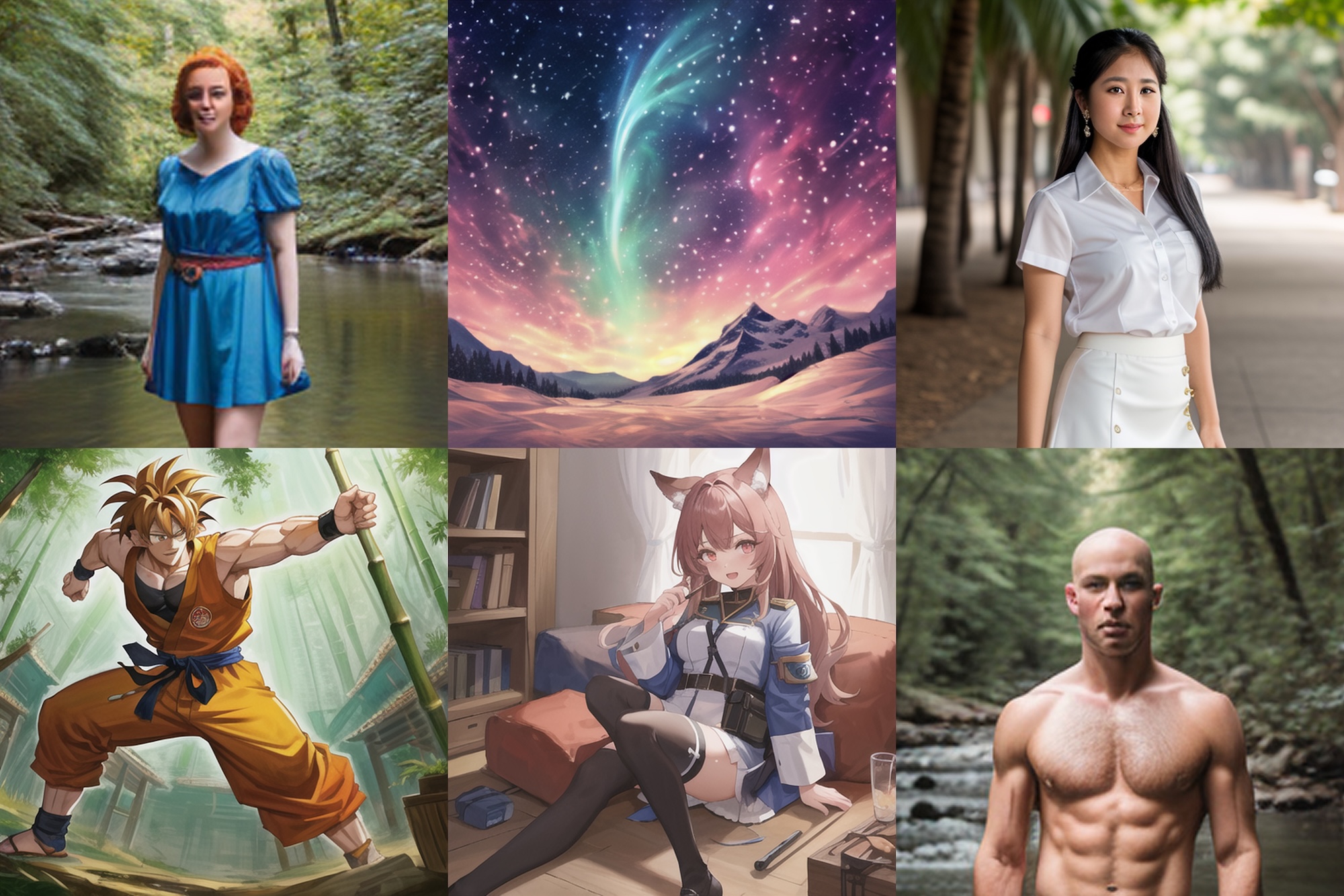}
    \caption{\textbf{Samples generated from the base model using text descriptions.} We observe distortions in the face, additional limbs, missing concepts, and inconsistent knowledge of characters and objects, all of which affect downstream tasks.}
    \label{fig:limit}
\end{figure}

\section{Limitations and Error Analysis}\label{sec:limitations}

Despite its training-free design, our method incurs computational cost. Runtime increases approximately linearly with the number of LoRA adapters when evaluated sequentially, as each adapter contributes a separate prediction. On an RTX A6000, a 512$\times$512 edit with five LoRAs requires 44s sequentially, compared to 27s for a text-guided edit using only the base objective. However, the noise predictions for each LoRA are computationally independent operations and can therefore be evaluated in a single batched forward pass, where runtime drops to \textbf{33s}, comparable to a single-LoRA edit.
Performance also depends on the quality and mutual alignment of LoRA adapters. Variability in publicly available LoRAs (\eg, inconsistent training data or rank) can introduce imbalance, occasionally biasing the composite output toward dominant adapters, an inherent issue of all training-free methods. Finally, generation quality remains bounded by the base model: artefacts such as duplicated limbs or entangled concepts (\eg, prompts such as ``raise arm'' that also cause a smile) stem from the base model's inherent priors, and are not explicitly addressed by our method (\cref{fig:limit}).
\section{Conclusion}
\label{sec:conclusion}
We present Concept Distillation Sampling (CDS), a novel, training-free framework for zero-shot, concept-based image editing. By integrating a robust, regularised distillation objective with a dynamic spatial concept-weighting mechanism, our method enables fine-grained, multi-concept edits. CDS successfully overcomes the linguistic limitations of text-only editing, maintaining strict subject fidelity even under complex transformations, such as simultaneous pose and semantic changes. Extensive quantitative and qualitative evaluations demonstrate SoTA performance across both text-guided and multi-concept edits.


%
%
\bibliographystyle{splncs04}

\begin{thebibliography}{10}
\providecommand{\url}[1]{\texttt{#1}}
\providecommand{\urlprefix}{URL }
\providecommand{\doi}[1]{https://doi.org/#1}

\bibitem{brooks2022instructpix2pix}
Brooks, T., Holynski, A., Efros, A.A.: Instructpix2pix: Learning to follow image editing instructions. In: {IEEE/CVF} Conference on Computer Vision and Pattern Recognition, {CVPR} 2023, Vancouver, BC, Canada, June 17-24, 2023. pp. 18392--18402. {IEEE} (2023). \doi{10.1109/CVPR52729.2023.01764}, \url{https://doi.org/10.1109/CVPR52729.2023.01764}

\bibitem{cao2023masactrl}
Cao, M., Wang, X., Qi, Z., Shan, Y., Qie, X., Zheng, Y.: Masactrl: Tuning-free mutual self-attention control for consistent image synthesis and editing. In: {IEEE/CVF} International Conference on Computer Vision, {ICCV} 2023, Paris, France, October 1-6, 2023. pp. 22503--22513. {IEEE} (2023). \doi{10.1109/ICCV51070.2023.02062}, \url{https://doi.org/10.1109/ICCV51070.2023.02062}

\bibitem{couairon2022diffedit}
Couairon, G., Verbeek, J., Schwenk, H., Cord, M.: Diffedit: Diffusion-based semantic image editing with mask guidance. In: The Eleventh International Conference on Learning Representations, {ICLR} 2023, Kigali, Rwanda, May 1-5, 2023. OpenReview.net (2023), \url{https://openreview.net/pdf?id=3lge0p5o-M-}

\bibitem{du2020compositional}
Du, Y., Li, S., Mordatch, I.: Compositional visual generation with energy based models. In: Larochelle, H., Ranzato, M., Hadsell, R., Balcan, M., Lin, H. (eds.) Advances in Neural Information Processing Systems 33: Annual Conference on Neural Information Processing Systems 2020, NeurIPS 2020, December 6-12, 2020, virtual (2020), \url{https://proceedings.neurips.cc/paper/2020/hash/49856ed476ad01fcff881d57e161d73f-Abstract.html}

\bibitem{feng2023trainingfree}
Feng, W., He, X., Fu, T., Jampani, V., Akula, A.R., Narayana, P., Basu, S., Wang, X.E., Wang, W.Y.: Training-free structured diffusion guidance for compositional text-to-image synthesis. In: The Eleventh International Conference on Learning Representations, {ICLR} 2023, Kigali, Rwanda, May 1-5, 2023. OpenReview.net (2023), \url{https://openreview.net/pdf?id=PUIqjT4rzq7}

\bibitem{foteinopoulou2025loratorio}
Foteinopoulou, N., Budvytis, I., Liwicki, S.: Loratorio: An intrinsic approach to lora skill composition. ArXiv preprint  \textbf{abs/2508.11624} (2025), \url{https://arxiv.org/abs/2508.11624}

\bibitem{gafni2022make}
Gafni, O., Polyak, A., Ashual, O., Sheynin, S., Parikh, D., Taigman, Y.: Make-a-scene: Scene-based text-to-image generation with human priors. In: European Conference on Computer Vision. pp. 89--106. Springer (2022)

\bibitem{Hertz_2023_ICCV}
Hertz, A., Aberman, K., Cohen{-}Or, D.: Delta denoising score. In: {IEEE/CVF} International Conference on Computer Vision, {ICCV} 2023, Paris, France, October 1-6, 2023. pp. 2328--2337. {IEEE} (2023). \doi{10.1109/ICCV51070.2023.00221}, \url{https://doi.org/10.1109/ICCV51070.2023.00221}

\bibitem{hu2022lora}
Hu, E.J., Shen, Y., Wallis, P., Allen{-}Zhu, Z., Li, Y., Wang, S., Wang, L., Chen, W.: Lora: Low-rank adaptation of large language models. In: The Tenth International Conference on Learning Representations, {ICLR} 2022, Virtual Event, April 25-29, 2022. OpenReview.net (2022), \url{https://openreview.net/forum?id=nZeVKeeFYf9}

\bibitem{huang2023lorahub}
Huang, C., Liu, Q., Lin, B.Y., Pang, T., Du, C., Lin, M.: Lorahub: Efficient cross-task generalization via dynamic lora composition. In: Conference on Language Modeling (2024)

\bibitem{huang2023composer}
Huang, L., Chen, D., Liu, Y., Shen, Y., Zhao, D., Zhou, J.: Composer: Creative and controllable image synthesis with composable conditions. In: Krause, A., Brunskill, E., Cho, K., Engelhardt, B., Sabato, S., Scarlett, J. (eds.) International Conference on Machine Learning, {ICML} 2023, 23-29 July 2023, Honolulu, Hawaii, {USA}. Proceedings of Machine Learning Research, vol.~202, pp. 13753--13773. {PMLR} (2023), \url{https://proceedings.mlr.press/v202/huang23b.html}

\bibitem{huggingface2024merge}
{Hugging Face}: Merging loras. \url{https://huggingface.co/docs/diffusers/en/using-diffusers/merge_loras} (2024), accessed: 2025-06-09

\bibitem{johnson2018image}
Johnson, J., Gupta, A., Fei{-}Fei, L.: Image generation from scene graphs. In: 2018 {IEEE} Conference on Computer Vision and Pattern Recognition, {CVPR} 2018, Salt Lake City, UT, USA, June 18-22, 2018. pp. 1219--1228. {IEEE} Computer Society (2018). \doi{10.1109/CVPR.2018.00133}, \url{http://openaccess.thecvf.com/content\_cvpr\_2018/html/Johnson\_Image\_Generation\_From\_CVPR\_2018\_paper.html}

\bibitem{kendall1939problem}
Kendall, M.G., Smith, B.B.: The problem of m rankings. The annals of mathematical statistics  \textbf{10}(3),  275--287 (1939)

\bibitem{kim2025dreamcatalyst}
Kim, J., Lee, S., Shin, J., Choi, J., Shim, H.: Dreamcatalyst: Fast and high-quality 3d editing via controlling editability and identity preservation. In: The Thirteenth International Conference on Learning Representations (2025), \url{https://openreview.net/forum?id=FA5ZAJlv96}

\bibitem{Koo:2024PDS}
Koo, J., Park, C., Sung, M.: Posterior distillation sampling. In: {IEEE/CVF} Conference on Computer Vision and Pattern Recognition, {CVPR} 2024, Seattle, WA, USA, June 16-22, 2024. pp. 13352--13361. {IEEE} (2024). \doi{10.1109/CVPR52733.2024.01268}, \url{https://doi.org/10.1109/CVPR52733.2024.01268}

\bibitem{kumari2023multi}
Kumari, N., Zhang, B., Zhang, R., Shechtman, E., Zhu, J.: Multi-concept customization of text-to-image diffusion. In: {IEEE/CVF} Conference on Computer Vision and Pattern Recognition, {CVPR} 2023, Vancouver, BC, Canada, June 17-24, 2023. pp. 1931--1941. {IEEE} (2023). \doi{10.1109/CVPR52729.2023.00192}, \url{https://doi.org/10.1109/CVPR52729.2023.00192}

\bibitem{labs2025flux1kontextflowmatching}
Labs, B.F., Batifol, S., Blattmann, A., Boesel, F., Consul, S., Diagne, C., Dockhorn, T., English, J., English, Z., Esser, P., Kulal, S., Lacey, K., Levi, Y., Li, C., Lorenz, D., Müller, J., Podell, D., Rombach, R., Saini, H., Sauer, A., Smith, L.: Flux.1 kontext: Flow matching for in-context image generation and editing in latent space (2025), \url{https://arxiv.org/abs/2506.15742}

\bibitem{li2023composing}
Li, S., Du, Y., Tenenbaum, J.B., Torralba, A., Mordatch, I.: Composing ensembles of pre-trained models via iterative consensus. In: The Eleventh International Conference on Learning Representations, {ICLR} 2023, Kigali, Rwanda, May 1-5, 2023. OpenReview.net (2023), \url{https://openreview.net/pdf?id=gmwDKo-4cY}

\bibitem{lin2023designbench}
Lin, K., Yang, Z., Li, L., Wang, J., Wang, L.: Designbench: Exploring and benchmarking dall-e 3 for imagining visual design. ArXiv preprint  \textbf{abs/2310.15144} (2023), \url{https://arxiv.org/abs/2310.15144}

\bibitem{liu2021learning}
Liu, N., Li, S., Du, Y., Tenenbaum, J., Torralba, A.: Learning to compose visual relations. In: Ranzato, M., Beygelzimer, A., Dauphin, Y.N., Liang, P., Vaughan, J.W. (eds.) Advances in Neural Information Processing Systems 34: Annual Conference on Neural Information Processing Systems 2021, NeurIPS 2021, December 6-14, 2021, virtual. pp. 23166--23178 (2021), \url{https://proceedings.neurips.cc/paper/2021/hash/c3008b2c6f5370b744850a98a95b73ad-Abstract.html}

\bibitem{lu2022dpm}
Lu, C., Zhou, Y., Bao, F., Chen, J., Li, C., Zhu, J.: Dpm-solver++: Fast solver for guided sampling of diffusion probabilistic models. ArXiv preprint  \textbf{abs/2211.01095} (2022), \url{https://arxiv.org/abs/2211.01095}

\bibitem{manor2026spanningvisualanalogyspace}
Manor, H., Gal, R., Maron, H., Michaeli, T., Chechik, G.: Spanning the visual analogy space with a weight basis of loras (2026), \url{https://arxiv.org/abs/2602.15727}

\bibitem{meral2024clora}
Meral, T.H.S., Simsar, E., Tombari, F., Yanardag, P.: Clora: A contrastive approach to compose multiple lora models (2024)

\bibitem{mokady2022null}
Mokady, R., Hertz, A., Aberman, K., Pritch, Y., Cohen{-}Or, D.: Null-text inversion for editing real images using guided diffusion models. In: {IEEE/CVF} Conference on Computer Vision and Pattern Recognition, {CVPR} 2023, Vancouver, BC, Canada, June 17-24, 2023. pp. 6038--6047. {IEEE} (2023). \doi{10.1109/CVPR52729.2023.00585}, \url{https://doi.org/10.1109/CVPR52729.2023.00585}

\bibitem{ouyang2025k}
Ouyang, Z., Li, Z., Hou, Q.: K-lora: Unlocking training-free fusion of any subject and style loras. ArXiv preprint  \textbf{abs/2502.18461} (2025), \url{https://arxiv.org/abs/2502.18461}

\bibitem{pooledreamfusion}
Poole, B., Jain, A., Barron, J.T., Mildenhall, B.: Dreamfusion: Text-to-3d using 2d diffusion. In: The Eleventh International Conference on Learning Representations, {ICLR} 2023, Kigali, Rwanda, May 1-5, 2023. OpenReview.net (2023), \url{https://openreview.net/pdf?id=FjNys5c7VyY}

\bibitem{Ramesh_2022_DALLE2}
Ramesh, A., Pavlov, P., Goh, G., Gray, S., Voss, C., Radford, A., Chen, M., Sutskever, I.: Hierarchical text-conditional image generation with clip latents (dall·e 2). vol. abs/2204.06125 (2022), \url{https://arxiv.org/abs/2204.06125}

\bibitem{rombach_high-resolution_2022}
Rombach, R., Blattmann, A., Lorenz, D., Esser, P., Ommer, B.: High-resolution image synthesis with latent diffusion models. In: Proceedings of the {IEEE}/{CVF} Conference on Computer Vision and Pattern Recognition. pp. 10684--10695 (2022)

\bibitem{roy2025multlfg}
Roy, A., Suin, M., Shah, K., Chellappa, R.: Multlfg: Training-free multi-lora composition using frequency-domain guidance. ArXiv preprint  \textbf{abs/2505.20525} (2025), \url{https://arxiv.org/abs/2505.20525}

\bibitem{ruiz2023dreambooth}
Ruiz, N., Li, Y., Jampani, V., Pritch, Y., Rubinstein, M., Aberman, K.: Dreambooth: Fine tuning text-to-image diffusion models for subject-driven generation. In: {IEEE/CVF} Conference on Computer Vision and Pattern Recognition, {CVPR} 2023, Vancouver, BC, Canada, June 17-24, 2023. pp. 22500--22510. {IEEE} (2023). \doi{10.1109/CVPR52729.2023.02155}, \url{https://doi.org/10.1109/CVPR52729.2023.02155}

\bibitem{ruiz2024hyperdreambooth}
Ruiz, N., Li, Y., Jampani, V., Wei, W., Hou, T., Pritch, Y., Wadhwa, N., Rubinstein, M., Aberman, K.: Hyperdreambooth: Hypernetworks for fast personalization of text-to-image models. In: {IEEE/CVF} Conference on Computer Vision and Pattern Recognition, {CVPR} 2024, Seattle, WA, USA, June 16-22, 2024. pp. 6527--6536. {IEEE} (2024). \doi{10.1109/CVPR52733.2024.00624}, \url{https://doi.org/10.1109/CVPR52733.2024.00624}

\bibitem{Saharia_2022_NeurIPS}
Saharia, C., Chan, W., Saxena, S., Li, L., Whang, J., Denton, E.L., Ghasemipour, S.K.S., Lopes, R.G., Ayan, B.K., Salimans, T., Ho, J., Fleet, D.J., Norouzi, M.: Photorealistic text-to-image diffusion models with deep language understanding. In: Koyejo, S., Mohamed, S., Agarwal, A., Belgrave, D., Cho, K., Oh, A. (eds.) Advances in Neural Information Processing Systems 35: Annual Conference on Neural Information Processing Systems 2022, NeurIPS 2022, New Orleans, LA, USA, November 28 - December 9, 2022 (2022), \url{http://papers.nips.cc/paper\_files/paper/2022/hash/ec795aeadae0b7d230fa35cbaf04c041-Abstract-Conference.html}

\bibitem{schuhmann2022laion5b}
Schuhmann, C., Beaumont, R., Vencu, R., Gordon, C., Wightman, R., Cherti, M., Coombes, T., Katta, A., Mullis, C., Wortsman, M., Schramowski, P., Kundurthy, S., Crowson, K., Schmidt, L., Kaczmarczyk, R., Jitsev, J.: {LAION-5B:} an open large-scale dataset for training next generation image-text models. In: Koyejo, S., Mohamed, S., Agarwal, A., Belgrave, D., Cho, K., Oh, A. (eds.) Advances in Neural Information Processing Systems 35: Annual Conference on Neural Information Processing Systems 2022, NeurIPS 2022, New Orleans, LA, USA, November 28 - December 9, 2022 (2022), \url{http://papers.nips.cc/paper\_files/paper/2022/hash/a1859debfb3b59d094f3504d5ebb6c25-Abstract-Datasets\_and\_Benchmarks.html}

\bibitem{shah2024ziplora}
Shah, V., Ruiz, N., Cole, F., Lu, E., Lazebnik, S., Li, Y., Jampani, V.: Ziplora: Any subject in any style by effectively merging loras. In: European Conference on Computer Vision. pp. 422--438. Springer (2024)

\bibitem{shenaj_lorarar_2024}
Shenaj, D., Bohdal, O., Ozay, M., Zanuttigh, P., Michieli, U.: {LoRA}.rar: {Learning} to {Merge} {LoRAs} via {Hypernetworks} for {Subject}-{Style} {Conditioned} {Image} {Generation} (2024), \url{https://arxiv.org/abs/2412.05148}

\bibitem{simsar2025loraclr}
Simsar, E., Hofmann, T., Tombari, F., Yanardag, P.: Loraclr: Contrastive adaptation for customization of diffusion models. In: Proceedings of the Computer Vision and Pattern Recognition Conference. pp. 13189--13198 (2025)

\bibitem{sohn2023styledrop}
Sohn, K., Ruiz, N., Lee, K., Chin, D.C., Blok, I., Chang, H., Barber, J., Jiang, L., Entis, G., Li, Y., et~al.: Styledrop: text-to-image generation in any style. In: Proceedings of the 37th International Conference on Neural Information Processing Systems. pp. 66860--66889 (2023)

\bibitem{song2021scorebased}
Song, Y., Sohl{-}Dickstein, J., Kingma, D.P., Kumar, A., Ermon, S., Poole, B.: Score-based generative modeling through stochastic differential equations. In: 9th International Conference on Learning Representations, {ICLR} 2021, Virtual Event, Austria, May 3-7, 2021. OpenReview.net (2021), \url{https://openreview.net/forum?id=PxTIG12RRHS}

\bibitem{wang2022diffusiondb}
Wang, Z.J., Montoya, E., Munechika, D., Yang, H., Hoover, B., Chau, D.H.: {D}iffusion{DB}: A large-scale prompt gallery dataset for text-to-image generative models. In: Rogers, A., Boyd-Graber, J., Okazaki, N. (eds.) Proceedings of the 61st Annual Meeting of the Association for Computational Linguistics (Volume 1: Long Papers). pp. 893--911. Association for Computational Linguistics, Toronto, Canada (2023). \doi{10.18653/v1/2023.acl-long.51}, \url{https://aclanthology.org/2023.acl-long.51}

\bibitem{zhongmulti2024}
Zhong, M., Wang, S., Lu, Y., Jiao, Y., Ouyang, S., Yu, D., Han, J., Chen, W., et~al.: Multi-lora composition for image generation. In: Transactions on Machine Learning Research (2024)

\bibitem{zhu_mole_2024}
Zhu, J., Chen, Y., Ding, M., Luo, P., Wang, L., Wang, J.: Mole: Enhancing human-centric text-to-image diffusion via mixture of low-rank experts. In: Globersons, A., Mackey, L., Belgrave, D., Fan, A., Paquet, U., Tomczak, J.M., Zhang, C. (eds.) Advances in Neural Information Processing Systems 38: Annual Conference on Neural Information Processing Systems 2024, NeurIPS 2024, Vancouver, BC, Canada, December 10 - 15, 2024 (2024), \url{http://papers.nips.cc/paper\_files/paper/2024/hash/3415a8f8127d5b0ceb7fd321180b1954-Abstract-Conference.html}

\bibitem{zou_cached_2025}
Zou, X., Shen, M., Bouganis, C.S., Zhao, Y.: Cached multi-lora composition for multi-concept image generation. In: 13th International Conference on Learning Representations (2025)

\end{thebibliography}

\clearpage
\setcounter{page}{1}
\appendix
\section{Sensitivity Analysis}
\label{app:sensitivity}
We conduct a sensitivity analysis of our method to the key hyperparameters defined in~\cref{sec:method}. For $\eta$, we evaluate the core CDS optimisation objective on 1000 random images from the InstructPix2Pix~\cite{brooks2022instructpix2pix} dataset, allowing for a direct comparison with baseline DDS. For patch size and sensitivity to temperature, we evaluate the full CDS framework on the $N=2-3$ anime subset of the ComposLoRA testbed.

\subsection{Sensitivity to Regularisation}
\begin{figure}[h]
    \centering
    \includegraphics[width=\linewidth]{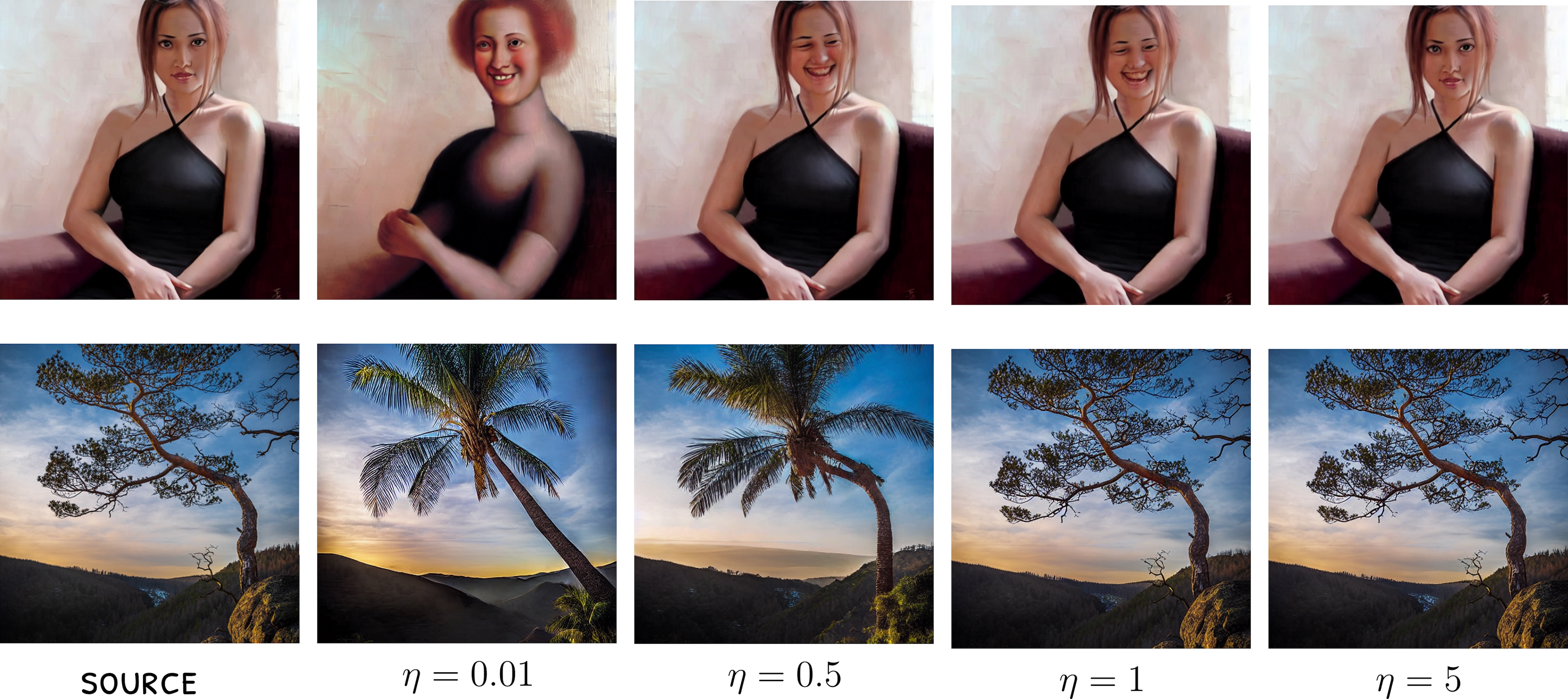}
    \caption{Sensitivity of the CDS objective to $\eta$. When the intended semantic change is spatially small, the method is highly robust to large $\eta$ values.}
    \label{fig:eta-samples}
\end{figure}
We evaluate the effect of the hyperparameter $\eta$ on the performance of the CDS optimisation objective.  
~\cref{tab:eta-sensitivity} reports CLIP and LPIPS scores across a wide range of $\eta$ values.  
Overall, we observe an inverse relationship between these metrics: a high CLIPScore indicates strong semantic alignment with the target prompt, while a very low LPIPS indicates minimal structural deviation from the source.  
~\cref{fig:eta-samples} further illustrates qualitative changes, showing that large $\eta$ values do not negatively affect the edit when the targeted edit area is small (\eg, ``add smile''), but they can overly constrain the optimisation and prevent edits over larger spatial areas (\eg, ``pine tree $\rightarrow$ palm tree''). As such, the optimal strength of the regularisation is highly dependent on the spatial footprint of the intended edit, with larger transformations requiring a lower $\eta$ and smaller, localised edits remaining robust under strong regularisation. We argue that because this regularisation depends on the semantic area of the edit rather than the base model's variance schedule, a simple user-guided parameter provides intuitive and reliable control over the editing process.

\begin{table}[h]
    \centering
    \caption{Sensitivity of the CDS objective to the $\eta$ hyperparameter, evaluated on InstructPix2Pix.}
    \label{tab:eta-sensitivity}
\begin{tabular}{c c c}
\hline
$\eta$ & CLIP $\uparrow$ & LPIPS $\downarrow$ \\
\hline
0.01 & $0.316 \pm 0.040$ & $0.407 \pm 0.151$ \\
0.05 & $0.308 \pm 0.042$ & $0.100 \pm 0.042$ \\
1    & $0.304 \pm 0.042$ & $0.090 \pm 0.047$ \\
5    & $0.299 \pm 0.044$ & $0.079 \pm 0.034$ \\
10   & $0.299 \pm 0.044$ & $0.089 \pm 0.035$ \\
\hline
\end{tabular}
\end{table}

\subsection{Sensitivity to Patch Size}

We further assess how patch size affects performance using a subset of the ComposLoRA testbed.  
Patch size controls the spatial granularity of the dynamic concept weighting, potentially impacting the method’s ability to capture and preserve structural composition.  
~\cref{tab:patch-size} shows that smaller patches generally yield better CLIP scores, suggesting that finer local evaluation of concept confidence improves the inclusion and isolation of specific semantic traits.

\begin{table}[h]
\centering
\caption{Effect of patch size on a subset of the ComposLoRA testbed. Although the method is relatively robust to patch size variations, a more localised approach (smaller patches) yields higher CLIPScores.}
\label{tab:patch-size}
        \begin{tabular}{llll}
        \toprule
                       & $N=2$                     & $N=3$                     & Avg. \\
            \midrule
                $2 \times 2$       & $\mathbf{0.340 \pm 0.030}$ & $\mathbf{0.361 \pm 0.022}$ & $\mathbf{0.351}$ \\
                $4 \times 4$       & $0.335 \pm 0.029$          & $0.356 \pm 0.022$          & $0.346$ \\
                $8 \times 8$       & $0.329 \pm 0.031$          & $0.349 \pm 0.020$          & $0.339$ \\
                $16 \times 16$     & $0.322 \pm 0.029$          & $0.341 \pm 0.022$          & $0.332$ \\
            \bottomrule
        \end{tabular}   
\end{table}

\subsection{Sensitivity to Temperature}
Finally, we analyse the temperature parameter $\tau$, which modulates the sharpness of the probability distribution during the patch-wise dynamic weighting process.  
~\cref{tab:temperature} indicates that CDS is robust to large changes in $\tau$; while performance does not collapse at higher temperatures, there is a clear improvement in CLIPScore at lower temperatures, indicating sharper spatial isolation and better concept inclusion, similar to the findings for patch size.  

\begin{table}[h]
\centering
\caption{Effect of $\tau$ on a subset of the ComposLoRA testbed. Lower temperatures positively improve CLIPScore by sharpening concept isolation.}
\label{tab:temperature}
        \begin{tabular}{llll}
        \toprule
                       & $N=2$                     & $N=3$                     & Avg. \\
            \midrule
                $\tau = 0.002$       & $\mathbf{0.340 \pm 0.030}$ & $\mathbf{0.361 \pm 0.022}$ & $\mathbf{0.351}$ \\
                $\tau = 0.01$        & $0.337 \pm 0.030$          & $0.360 \pm 0.022$          & $0.349$ \\
                $\tau = 0.1$         & $0.333 \pm 0.029$          & $0.360 \pm 0.022$          & $0.347$ \\
                $\tau = 1$           & $0.329 \pm 0.029$          & $0.359 \pm 0.022$          & $0.344$ \\
            \bottomrule
        \end{tabular}   
\end{table}

\section{Backbone Generalisation: SDXL}
\label{app:sdxl}

\begin{figure}[h]
\centering
\setlength{\tabcolsep}{2pt}
\renewcommand{\arraystretch}{1.0}
\begin{tabular}{@{}ccc@{}}
\includegraphics[width=0.28\linewidth]{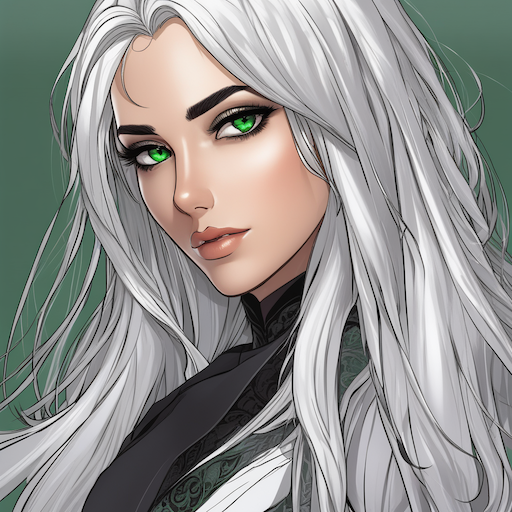} &
\includegraphics[width=0.28\linewidth]{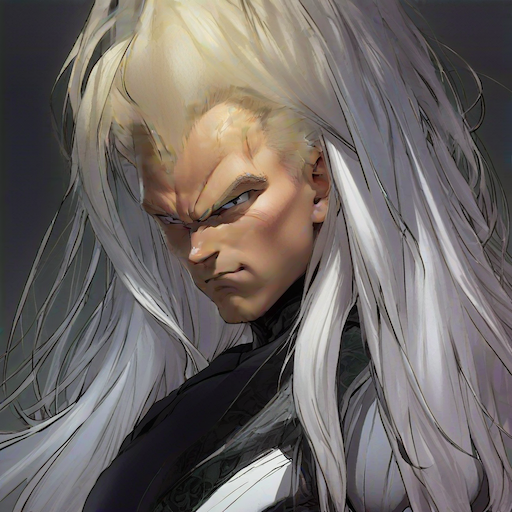} &
\includegraphics[width=0.28\linewidth]{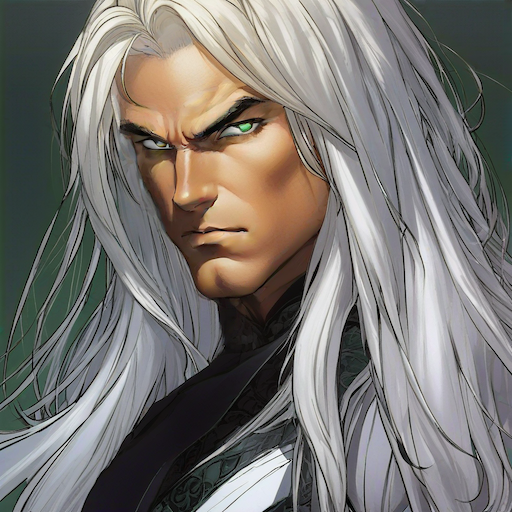} \\
{\scriptsize Source} & {\scriptsize Merge} & {\scriptsize CDS (Ours)} \\
\end{tabular}
\caption{Sample character + style change using SDXL. Merge collapses the target identity; CDS preserves source structure while introducing both concepts.}
\label{fig:sdxl-preview}
\end{figure}

While the main results use Stable Diffusion v1.5 for parity with prior multi-LoRA composition baselines, the CDS formulation itself is architecture-agnostic: the ordered-timestep schedule and the LoRA-vs-base divergence proxy require only access to a noise (or velocity) head and therefore transfer to newer DiT- and flow-matching backbones. To demonstrate this, we report an evaluation on SDXL at $N{=}2$ in~\cref{tab:sdxl} and~\cref{fig:sdxl-preview}, using the same ComposLoRA-style protocol as the main experiments with publicly available SDXL LoRAs. CDS preserves the source identity considerably more tightly than the Merge baseline (LPIPS $0.21$ vs.\ $0.38$) while achieving a higher CLIPScore ($30.40$ vs.\ $27.28$), confirming that the gains observed under SD\,v1.5 are not backbone-specific. We leave a full multi-$N$ evaluation on SDXL and a port to flow-matching models (\eg, SD3, Flux) to future work.

\begin{table}[h]
\centering
\caption{ SDXL results at $N=2$ on a ComposLoRA-style protocol. CDS retains its identity-preservation and semantic-alignment advantages on a stronger backbone.}
\label{tab:sdxl}
\begin{tabular}{lcc}
\toprule
            & CLIPScore$\uparrow$ & LPIPS$\downarrow$ \\
\midrule
Merge       & $0.273$             & $0.38$ \\
CDS (Ours)  & $\mathbf{0.304}$    & $\mathbf{0.21}$ \\
\bottomrule
\end{tabular}
\end{table}

\section{Qualitative Examples}
\label{app:qualitative}

\begin{figure*}[ht]
    \centering
    \includegraphics[width=\linewidth]{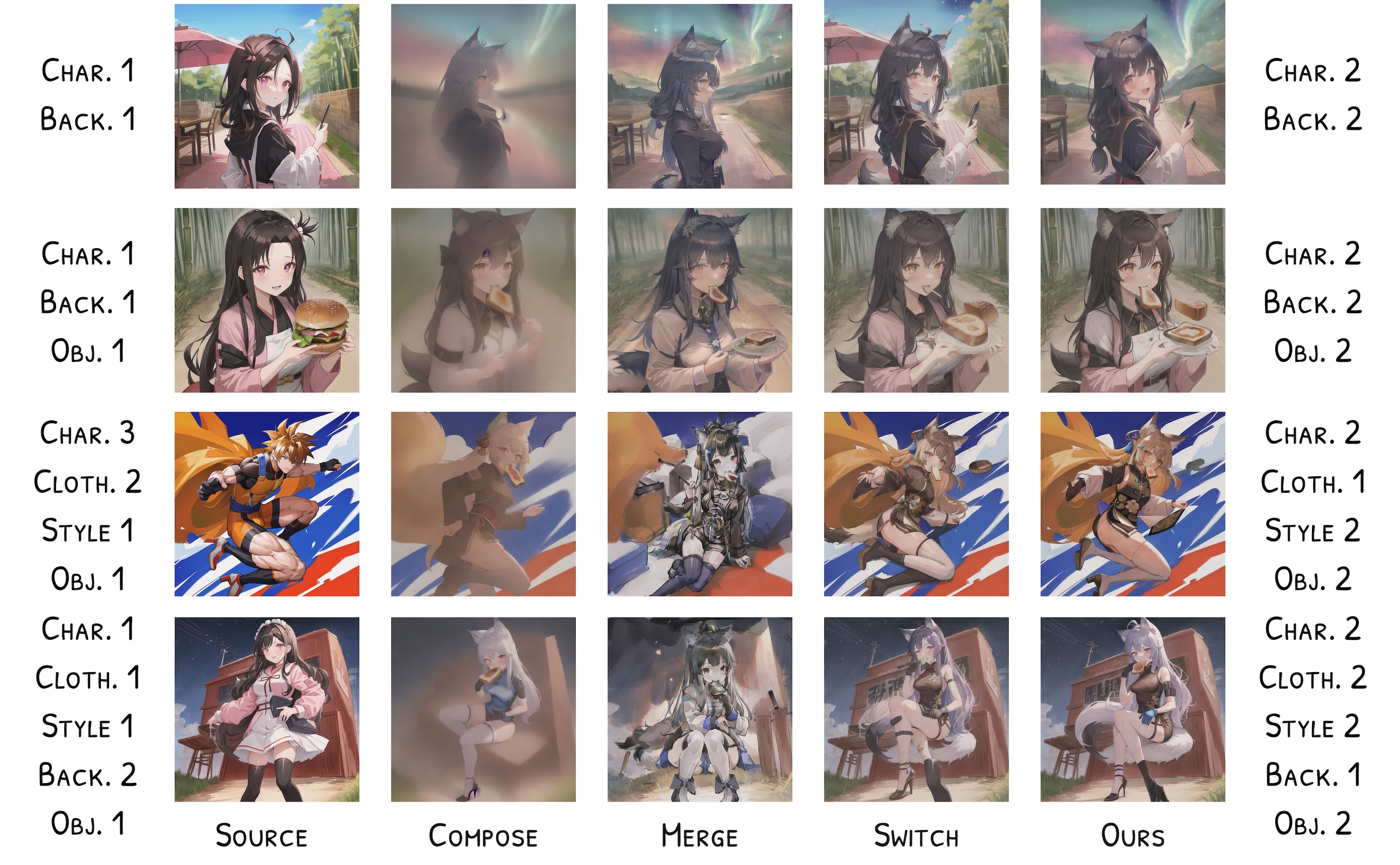}
    \caption{Examples of multi-concept edits on the anime subset of the ComposLoRA testbed for various $N$.}\label{fig:qual_anime}
\end{figure*}
As discussed in~\cref{sec:results}, we select existing multi-LoRA strategies that have publicly available code we
can adopt to image editing; this is achieved by combining our optimisation objective described in~\cref{sec:optimised_dds} and switching out the dynamic concept weighting. We exclude works such as LoRWeB~\cite{manor2026spanningvisualanalogyspace}, as the weights and dataset are not available; therefore, to our knowledge, replicating the results is not currently feasible. Furthermore, LoRWeB is limited to a fusion of two concepts and is not suitable for the Multi-LoRA setting.

\begin{figure*}[t]
    \centering
    \includegraphics[width=\linewidth]{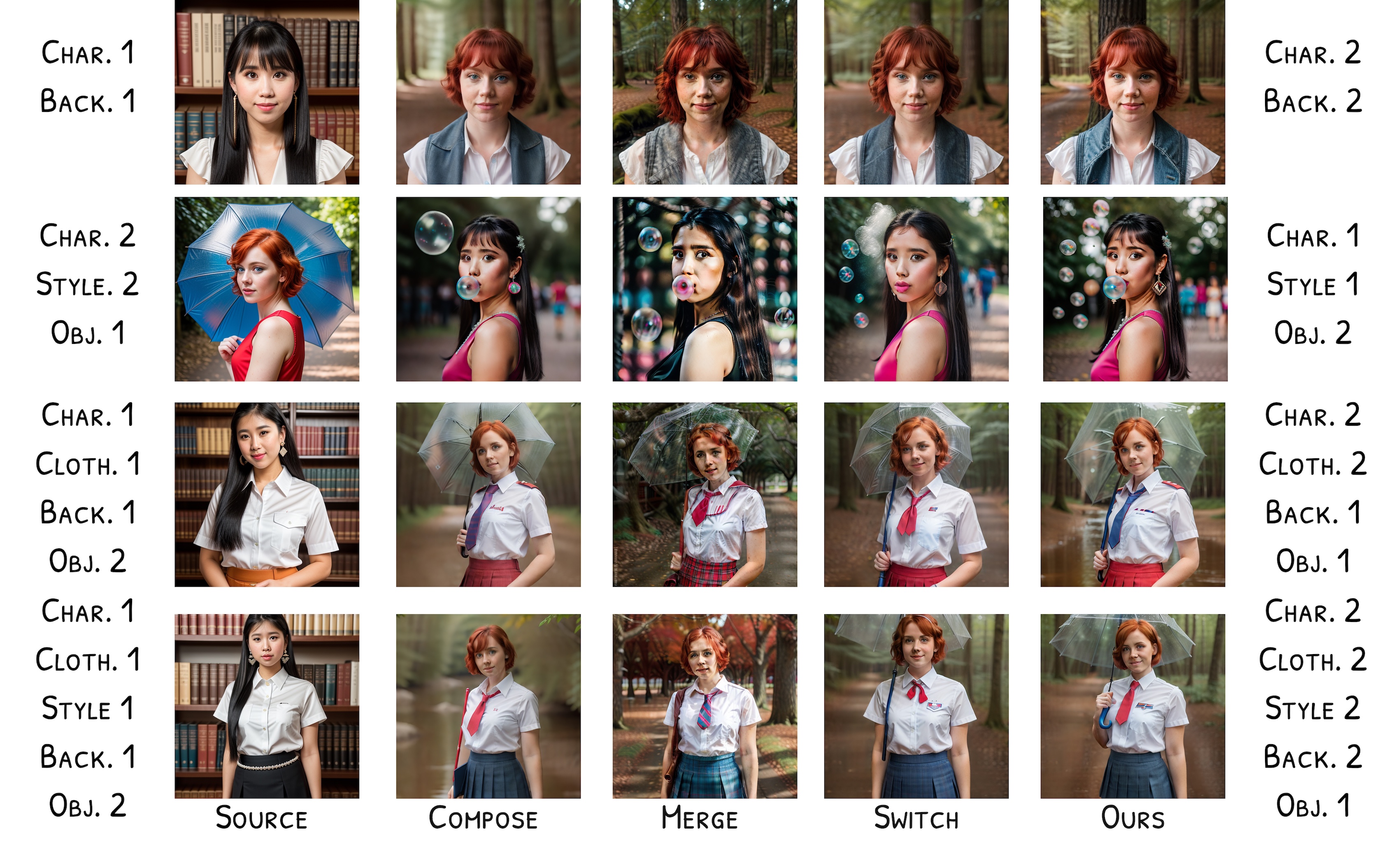}
    \caption{Examples of multi-concept edits on the reality subset of the ComposLoRA testbed for various $N$.}\label{fig:qual_reality}
\end{figure*}

In~\cref{fig:qual_anime} and~\ref{fig:qual_reality}, we observe several qualitative trends that reinforce the motivations and findings presented in the main paper. First, the source images—generated using a naïve text-only baseline—frequently omit key elements, underscoring the necessity of LoRA-driven concept composition even in ostensibly simple scenarios, as discussed in~\cref{sec:limitations}. However, when editing these scenes, CDS exhibits the lowest degree of concept erasure: all intended elements are consistently retained or added in the final outputs, aligning perfectly with the framework’s goal of preserving identity throughout the editing process. 

While all methods perform comparably for smaller compositions (\eg, $N = 2$), increasing the number of concepts makes the spatial interference problem significantly more challenging, which highlights our method's advantage. Specifically, we observe that \textit{Merge} introduces cumulative visual artefacts, whereas \textit{Switch} progressively drops concepts entirely as $N$ grows. In contrast, CDS remains highly stable, maintaining coherent structural integrity and concept fidelity even in demanding multi-LoRA scenarios. This reflects the robustness of unifying a regularised distillation objective with dynamic spatial concept weighting.

\section{Qualitative Results}
\label{app:qualitative-results}
\begin{figure}[h]
    \centering
    \begin{subfigure}[b]{0.45\linewidth}
        \centering
        \includegraphics[width=0.9\linewidth]{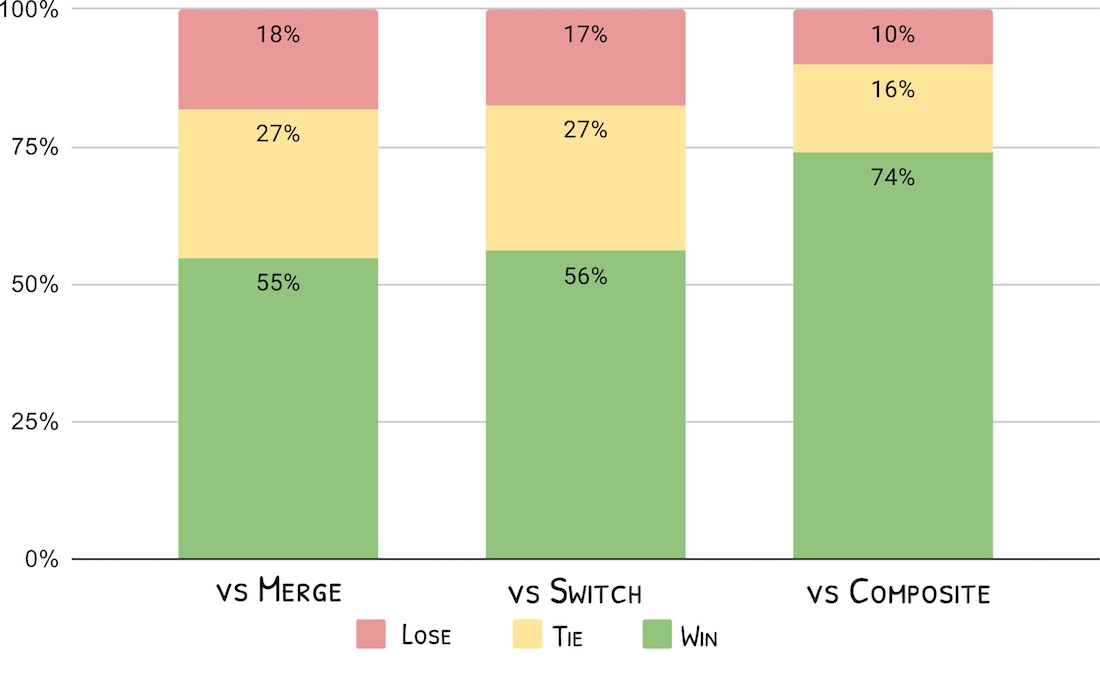}
        \caption{Anime Subset.}
        \label{fig:gptv_anime}
    \end{subfigure}  
    \begin{subfigure}[b]{0.45\linewidth}
        \centering
        \includegraphics[width=0.9\linewidth]{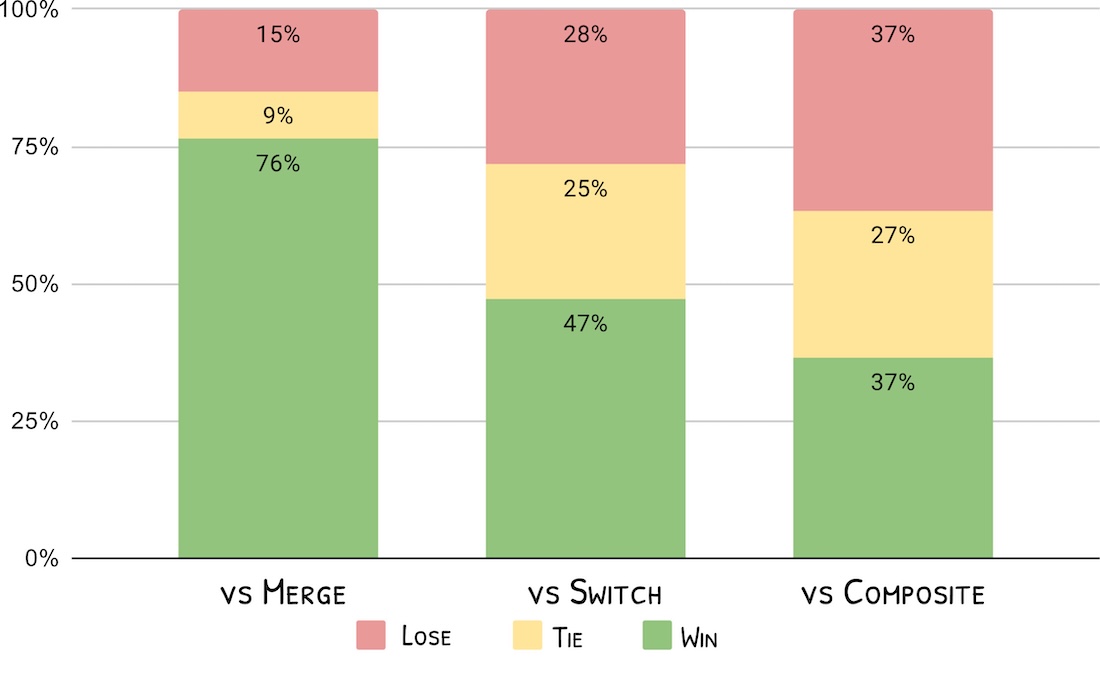}
        \caption{Reality Subset.}
        \label{fig:gpt4v_real}
    \end{subfigure}
    \caption{\textbf{Comparison of our CDS framework against previous LoRA-composition methods, evaluated using GPT-4V.} Our method achieves consistently higher pairwise win rates across both subsets.}
    \label{fig:gptv_reality}
\end{figure}

In addition to the aggregate results in~\cref{sec:gpt4v}, we provide a breakdown of GPT-4V evaluation win rates separated by the stylistic subsets of the ComposLoRA testbed in~\cref{fig:gptv_reality}. We also provide a detailed breakdown of the Image Quality and Composition Quality scores for the two subsets in~\cref{tab:gptv_table_app}.

\begin{table*}[t]
    \centering
    \caption{\textbf{Comparison of Concept Distillation Sampling (CDS) against previous LoRA-composition methods, evaluated using GPT-4V.} Our method consistently achieves the highest total Image and Composition Quality scores.}
    \label{tab:gptv_table_app}
    \begin{subtable}[t]{\textwidth}
    \resizebox{\textwidth}{!}{%
    \begin{tabular}{lllllllllll}
    \hline
                       & \multicolumn{2}{c}{\textbf{$N=2$}} & \multicolumn{2}{c}{\textbf{$N=3$}} & \multicolumn{2}{c}{\textbf{$N=4$}} & \multicolumn{2}{c}{\textbf{$N=5$}} & \multicolumn{2}{l}{Total} \\ \cline{2-11} 
                       & \multicolumn{1}{c}{Image Q.} & \multicolumn{1}{c}{Composition Q.} & \multicolumn{1}{c}{Image Q.} & \multicolumn{1}{c}{Composition Q.} & \multicolumn{1}{c}{Image Q.} & \multicolumn{1}{c}{Composition Q.} & \multicolumn{1}{c}{Image Q.} & \multicolumn{1}{c}{Composition Q.} & \multicolumn{1}{c}{Image Q.} & \multicolumn{1}{c}{Composition Q.} \\ \hline
    \textbf{Merge}     & 8.42 & 7.83 & 8.27 & 7.95 & 8.62 & 7.59 & 8.23 & 6.35 & 8.39 & 7.43 \\
    \textbf{Switch}    & 8.46 & 8.21 & 8.57 & 7.91 & 8.74 & 6.97 & 8.69 & 6.6  & \uline{8.62} & \uline{7.42} \\
    \textbf{Composite} & 7.94 & 7.46 & 8.02 & 7.43 & 7.94 & 6.94 & 7.84 & 6.02 & 7.94 & 6.96 \\
    \textbf{Ours}      & 8.89 & 8.85 & 8.83 & 8.47 & 8.88 & 7.86 & 8.87 & 7.15 & \textbf{8.87} & \textbf{8.08} \\ \hline
    \end{tabular}%
    }
    \caption{Anime Subset}
    \end{subtable}

    \begin{subtable}[t]{\textwidth}
    \resizebox{\textwidth}{!}{%
    \begin{tabular}{lllllllllll}
    \hline
                       & \multicolumn{2}{c}{\textbf{$N=2$}} & \multicolumn{2}{c}{\textbf{$N=3$}} & \multicolumn{2}{c}{\textbf{$N=4$}} & \multicolumn{2}{c}{\textbf{$N=5$}} & \multicolumn{2}{l}{Total} \\ \cline{2-11} 
                       & \multicolumn{1}{c}{Image Q.} & \multicolumn{1}{c}{Composition Q.} & \multicolumn{1}{c}{Image Q.} & \multicolumn{1}{c}{Composition Q.} & \multicolumn{1}{c}{Image Q.} & \multicolumn{1}{c}{Composition Q.} & \multicolumn{1}{c}{Image Q.} & \multicolumn{1}{c}{Composition Q.} & \multicolumn{1}{c}{Image Q.} & \multicolumn{1}{c}{Composition Q.} \\ \hline
    \textbf{Merge}     & 7.56 & 7.94 & 7.38 & 7.45 & 7.56 & 6.89 & 7.33 & 6.5  & 7.46 & 7.20 \\
    \textbf{Switch}    & 8.74 & 8.37 & 8.7  & 8.23 & 8.39 & 7.16 & 8.1  & 7.0  & 8.48 & 7.69 \\
    \textbf{Composite} & 8.75 & 8.65 & 8.73 & 8.44 & 8.73 & 8.02 & 8.67 & 7.9  & \uline{8.72} & \uline{8.25} \\
    \textbf{Ours}      & 8.87 & 8.81 & 8.78 & 8.5  & 8.74 & 8.21 & 8.85 & 8.11 & \textbf{8.81} & \textbf{8.41} \\ \hline
    \end{tabular}%
    }
    \caption{Reality Subset}
    \end{subtable}
\end{table*}

\section{GPT-4V Evaluation Interface}
To complement our quantitative metrics and human studies, we conduct a large-scale qualitative evaluation using GPT-4V, following and adapting the methodology of~\cite{zhongmulti2024}. Unlike prior work focused on text-to-image generation, our task involves evaluating zero-shot image editing quality. Therefore, we modify the evaluation prompt to reflect the specific dual-goals of editing—namely, preserving structural source content while achieving complex, concept-driven target modifications.

For each test case, GPT-4V is presented with: (a) two edited outputs (from our method and a baseline), and (b) a structured description of the source composition and the intended target composition.

GPT-4V is asked to evaluate the two edited images across two dimensions: Image Quality (\ie, assessing visual fidelity, absence of generative artefacts, and overall aesthetic) and Composition Quality (\ie, assessing how well the edit reflects the intended multi-concept transformation while preserving identity and structure).

Each pairwise comparison is conducted twice with the image order reversed to mitigate positional bias. Scores are averaged across both runs. The evaluation prompt is detailed in~\cref{lst:gpt4v_prompt}.

\section{ComposLoRA Dataset Details}
\label{appendix:composlora}

\cref{tab:composlora_details} provides the concepts, trigger words and categories for the LoRAs used in the ComposLoRA testbed, as introduced by \cite{zhongmulti2024}.

\begin{table}[ht]
\centering
\small
\caption{Detailed descriptions of each LoRA in the ComposLoRA~\cite{zhongmulti2024} Testbed.}
\label{tab:composlora_details}
\begin{tabular}{llp{5cm}l}
\toprule
\textbf{Category} & \textbf{LoRA Name} & \textbf{Trigger Words} & \textbf{Source} \\ \midrule
\multicolumn{4}{c}{\textit{Anime Style Subset}} \\ \midrule
Character & Kamado Nezuko & kamado nezuko, black hair, pink eyes, forehead & \href{https://civitai.com/models/20340}{Link} \\ 
Character & Texas the Omertosa & omertosa, 1girl, wolf ears, long hair & \href{https://civitai.com/models/112610}{Link} \\ 
Character & Son Goku & son goku, spiked hair, muscular male, wristband & \href{https://civitai.com/models/13175}{Link} \\
Clothing & Garreg Mach Uniform & gmuniform, blue thighhighs, long sleeves & \href{https://civitai.com/models/30521}{Link} \\
Clothing & Zero Suit (Metroid) & zero suit, blue gloves, high heels & \href{https://civitai.com/models/15729}{Link} \\
Style & Hand-drawn Style & lineart, hand-drawn style & \href{https://civitai.com/models/16014}{Link} \\ 
Style & Chinese Ink Wash & shuimobysim, traditional chinese ink painting & \href{https://civitai.com/models/64344}{Link} \\ 
Background & Bamboolight & bamboolight, outdoors, bamboo & \href{https://civitai.com/models/21262}{Link} \\ 
Background & Auroral Background & auroral, starry sky, outdoors & \href{https://civitai.com/models/12548}{Link} \\
Object & Huge Burger & two-handed burger, holding a huge burger with both hands & \href{https://civitai.com/models/113197}{Link} \\
Object & Toast & toast, toast in mouth & \href{https://civitai.com/models/126989}{Link} \\
\midrule
 \multicolumn{4}{c}{\textit{Realistic Style Subset}} \\ \midrule
Character & IU (Lee Ji Eun) & iu1, long straight black hair, hazel eyes, diamond stud earrings & \href{https://civitai.com/models/50930}{Link} \\ 
Character & Scarlett Johansson & scarlett, short red hair, blue eyes & \href{https://civitai.com/models/14413}{Link} \\ 
Character & The Rock & th3r0ck with no hair, muscular male, serious look on his face & \href{https://civitai.com/models/12626}{Link} \\ 
Clothing & Thai Univ. Uniform & mahalaiuniform, white shirt short sleeves, black pencil skirt & \href{https://civitai.com/models/78891}{Link} \\
Clothing & School Dress & school uniform, white shirt, red tie, blue pleated microskirt & \href{https://civitai.com/models/11225}{Link} \\ 
Style & Japanese Film Color & film overlay, film grain & \href{https://civitai.com/models/101291}{Link} \\ 
Style & Bright Style & bright lighting & \href{https://civitai.com/models/93988}{Link} \\ 
Background & Library Bookshelf & lib\_bg, library bookshelf & \href{https://civitai.com/models/113054}{Link} \\ 
Background & Forest Background & slg, river, forest & \href{https://civitai.com/models/125350}{Link} \\ 
Object & Umbrella & transparent umbrella & \href{https://civitai.com/models/112260}{Link} \\
Object & Bubble Gum & blow bubble gum & \href{https://civitai.com/models/122708}{Link} \\ \bottomrule
\end{tabular}
\end{table}

\clearpage

\section{Human Evaluation}
\label{app:human}
For the human qualitative evaluation, we asked three human experts to rank images produced by the four methods. For each combination, we provide a set of reference images and the output of the anonymised methods. Samples of the instructions and survey, as shown to human experts, are presented below. Kendall's coefficient of concordance~\cite{kendall1939problem} is used as a metric of inter-annotator agreement for the ranking, with a value of $0.68$ showing substantial agreement.

\begin{lstlisting}[caption={GPT-4V Evaluation Prompt}, label={lst:gpt4v_prompt}]
I need assistance in evaluating image editing methods. You will see two images:

1. Edited image from Method 1
2. Edited image from Method 2

SOURCE COMPOSITION:
<source_elements>

TARGET COMPOSITION (what the edited images should achieve):
<target_elements>

Please evaluate both edited images (Images 1 and 2) on the following two dimensions:

1. IMAGE QUALITY (0-10, 0.5 increments):
- Deduct 3 points for each deformity (extra limbs, distorted faces, incorrect proportions)
- Deduct 2 points for noticeable issues with texture, lighting, or color
- Deduct 1 point for minor flaws or imperfections
- Additional deductions can be made for any issues affecting the overall aesthetic or clarity of the image.

2. COMPOSITION QUALITY (0-10, 0.5 increments):
- Deduct 3 points if any target element is missing or incorrectly depicted
- Deduct 1 point for each missing feature within an element
- Deduct 1 point for minor inconsistencies or lack of harmony
- Additional deductions can be made for compositions that lack coherence, creativity, or realism.

Please format the evaluation as follows:

For Image 1:
[Explanation of evaluation]

For Image 2:
[Explanation of evaluation]

Scores:
Image 1: Composition Quality: [score]/10, Image Quality: [score]/10
Image 2: Composition Quality: [score]/10, Image Quality: [score]/10

Based on the above guidelines, help me to conduct a step-by-step comparative evaluation of the given images. The scoring should follow two principles:
1. Please evaluate critically.
2. Try not to let the two models end in a tie on both dimensions.
\end{lstlisting}

\begin{figure*}[t]
  \centering
  \includegraphics[width=0.95\linewidth]{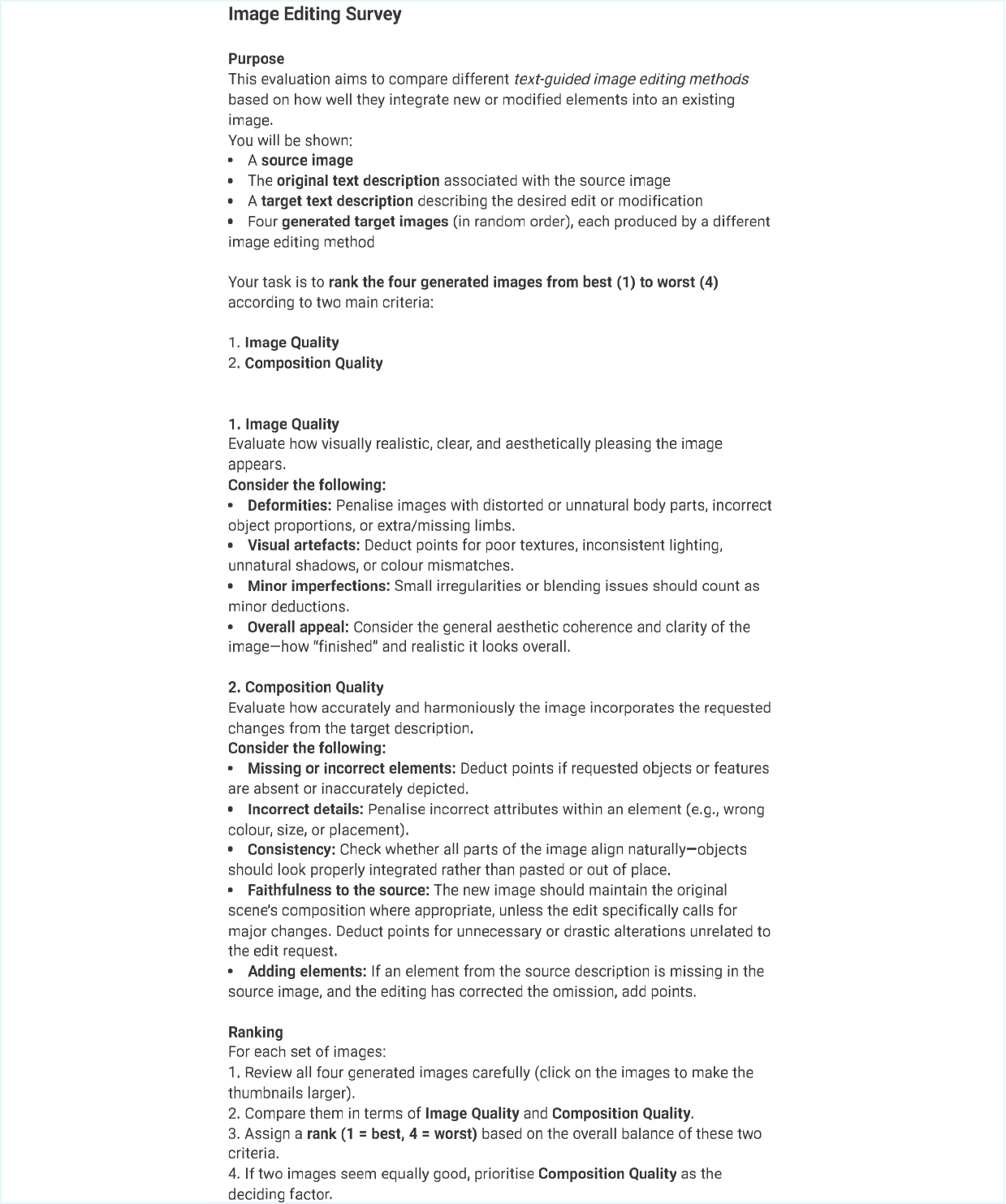}
  \caption{Human Instructions for the Multi-Concept Editing task.}
  \label{fig:instructions}
\end{figure*}

\begin{figure*}[t]
  \centering
  \includegraphics[width=0.95\linewidth]{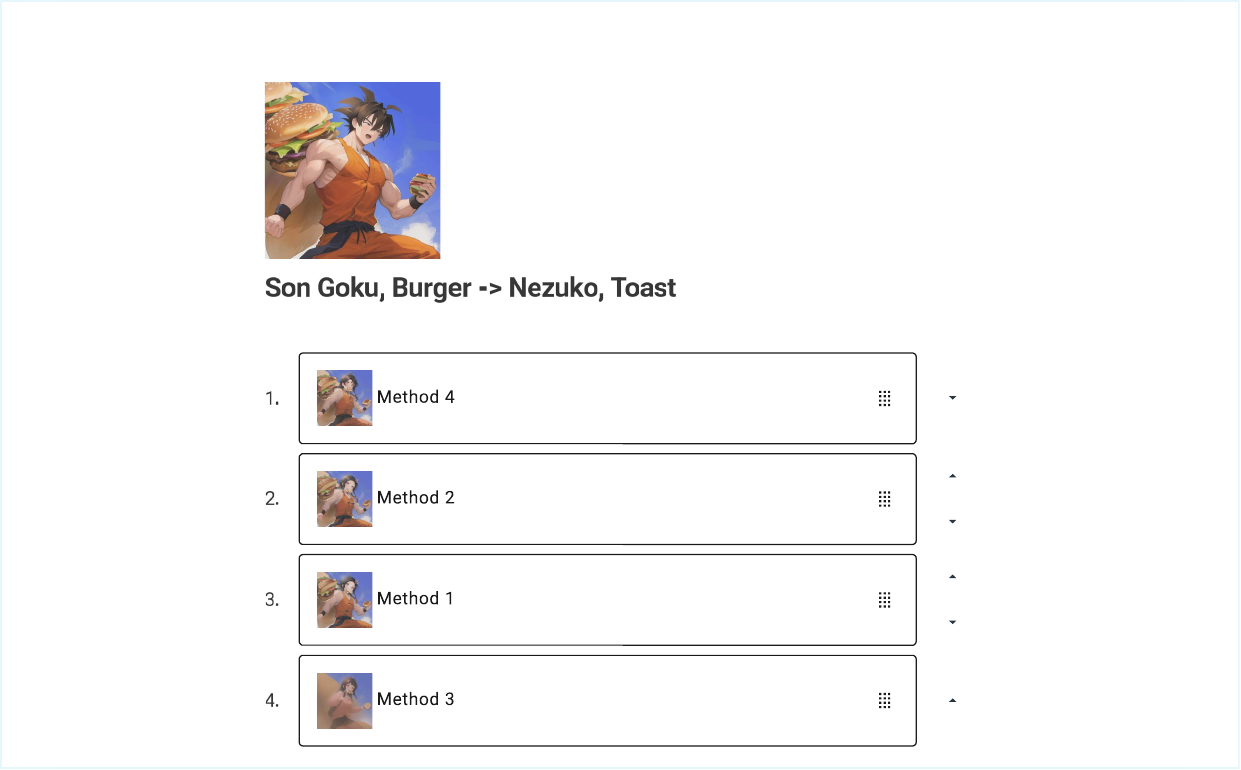}
  \caption{Example Human Evaluation Interface.}
  \label{fig:human_interface}
\end{figure*}

\end{document}